%% file: arxiv.tex
\PassOptionsToPackage{table,xcdraw}{xcolor}
\documentclass[10pt,twocolumn,letterpaper]{article}

\usepackage[pagenumbers]{iccv} 


\definecolor{iccvblue}{rgb}{0.21,0.49,0.74}
\usepackage[pagebackref,breaklinks,colorlinks,allcolors=iccvblue]{hyperref}

\usepackage{graphicx}
\usepackage{amsmath}
\usepackage{amssymb}
\usepackage{booktabs}
\usepackage{times}
\usepackage{epsfig}
\usepackage{multirow}
\usepackage{color}
\usepackage{arydshln}
\usepackage{float}
\usepackage{caption}
\usepackage{pifont}
\newcommand{\cmark}{\ding{51}}%
\newcommand{\xmark}{\ding{55}}%

\usepackage[pagebackref,breaklinks,colorlinks]{hyperref}

\usepackage[capitalize]{cleveref}
\crefname{section}{Sec.}{Secs.}
\Crefname{section}{Section}{Sections}
\Crefname{table}{Table}{Tables}
\crefname{table}{Tab.}{Tabs.}

\definecolor{best}{HTML}{bcf9bb}

\definecolor{secondbest}{HTML}{ffacac}

\def\paperID{239} 
\def\confName{ICCV}
\def\confYear{2025}

\title{Time-Aware Auto White Balance in Mobile Photography}

\author{\\[-20pt] 
Mahmoud Afifi\textsuperscript{*} \hspace{15pt} Luxi Zhao\textsuperscript{*} \hspace{15pt} Abhijith Punnappurath \\
Mohammed A. Abdelsalam \hspace{15pt} Ran Zhang \hspace{15pt} Michael S. Brown \\
AI Center–Toronto, Samsung Electronics \\
{\tt\small \{m.afifi1, lucy.zhao, abhijith.p, m.abdelsalam, ran.zhang, michael.b1\}@samsung.com}\\[-20pt]  
}

\begin{document}

\twocolumn[{%
\renewcommand\twocolumn[1][]{#1}%
\maketitle
\begin{center}
\includegraphics[width=\textwidth]{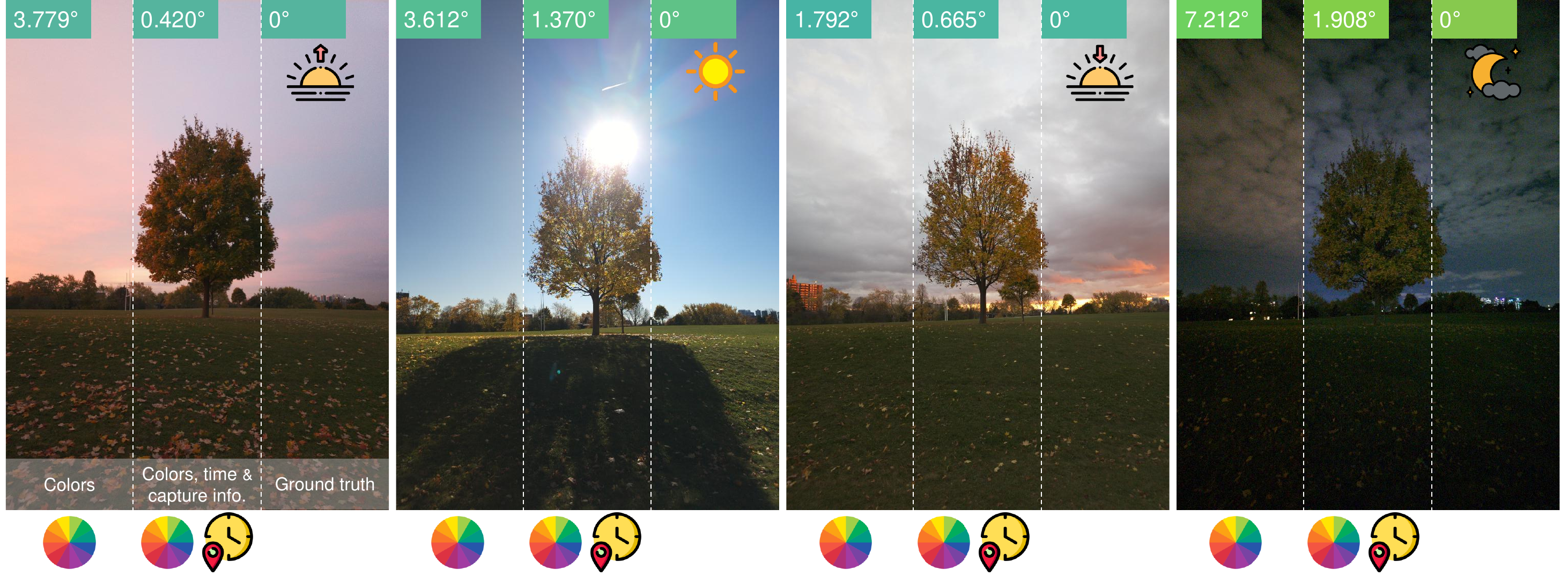}
\vspace{-6mm}
\captionof{figure}{
The time of day influences scene illumination, making it a valuable cue for improving illuminant estimation. Shown are white-balanced results (gamma-corrected for visualization) using our method with (1) colors only, (2) colors plus contextual (timestamp and geolocation) and capture data, and (3) ground truth (from a color chart). Angular errors show improvements when time-capture information is used. \label{fig:teaser}}\vspace{-2mm}
\end{center}%
}]

\maketitle
\def\thefootnote{*}\footnotetext{Equal contribution.}
\def\thefootnote{\arabic{footnote}}  

\begin{abstract}
Cameras rely on auto white balance (AWB) to correct undesirable color casts caused by scene illumination and the camera’s spectral sensitivity. This is typically achieved using an illuminant estimator that determines the global color cast solely from the color information in the camera's raw sensor image. Mobile devices provide valuable additional metadata---such as capture timestamp and geolocation---that offers strong contextual clues to help narrow down the possible illumination solutions. This paper proposes a lightweight illuminant estimation method that incorporates such contextual metadata, along with additional capture information and image colors, into a lightweight model ($\sim$5K parameters), achieving promising results, matching or surpassing larger models. To validate our method, we introduce a dataset of 3,224 smartphone images with contextual metadata collected at various times of day and under diverse lighting conditions. The dataset includes ground-truth illuminant colors, determined using a color chart, and user-preferred illuminants validated through a user study, providing a comprehensive benchmark for AWB evaluation.
\end{abstract}

\section{Introduction and related work}
\label{sec:intro}

Color constancy refers to the ability of the human visual system to maintain stable object colors despite variations in lighting conditions by leveraging contextual cues within the scene \cite{hansen2007effects, gegenfurtner2024color, maloney2002illuminant}. Cameras approximate this effect using auto white balance (AWB) correction, which aims to partially neutralize color casts introduced by scene illumination and the camera’s spectral sensitivity \cite{afifi2020deep}. AWB first estimates the illumination color as an RGB vector in the camera's raw color space. The raw image is then corrected by scaling its color channels according to the estimated illumination, typically under the assumption of a single global light source \cite{agarwal2006overview, gijsenij2011computational}.

Conventional illuminant estimation methods primarily rely on image colors, either by directly processing the raw image (e.g., \cite{GE, wGE, BMVC1, shi2016deep, CLASSIFICATION, C4, BoCF, QUASI-CC}) or by analyzing color histograms (e.g., \cite{CCC, SIIE, C5, afifi2025optimizing}). These methods can be broadly categorized into two groups: (1) classical statistical-based methods (e.g., \cite{GW, SoG, GE, wGE, MSGP, GI}), which estimate the illuminant color based on image statistics, and (2) learning-based methods (e.g., \cite{GAMUT, CCC, C5, TLCC, CWCC}), which map image colors to their corresponding scene illuminant through data-driven models.

While image colors are a key cue for estimating the scene’s illuminant, mobile devices provide an opportunity to integrate additional contextual information. For instance, the device’s location, along with the date and time, offer valuable cues about outdoor lighting conditions (e.g., sunrise, noon, sunset), thereby improving illuminant estimation for outdoor scenes (see Fig.~\ref{fig:teaser}).

\begin{figure}[t]
\centering
\includegraphics[width=0.98\linewidth]{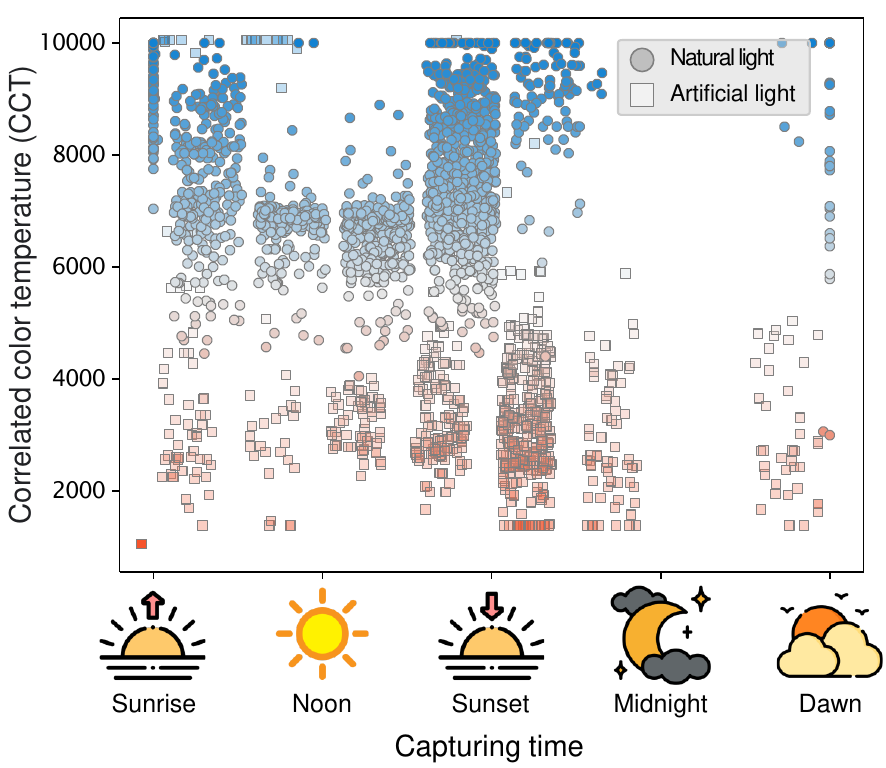}
\caption{The time of day at which an image is captured provides valuable information about the possible range of illuminant colors in outdoor scenes. The figure presents the correlated color temperature (CCT) of illuminant colors in our dataset (Sec.~\ref{sec:dataset}) for images captured at various times throughout the day and night. As shown, excluding images taken under artificial light, those captured at noon, for example, exhibit a different range of illuminant CCTs compared to images captured during sunset or sunrise.\label{fig:prob_distribution}}
\vspace{-2mm}
\end{figure}

Intuitively, knowing the time of day when an outdoor scene is captured can help estimate the lighting conditions. Figure~\ref{fig:prob_distribution} further illustrates that the time of day, derived from contextual metadata (i.e., timestamp and geolocation) readily available on mobile devices, provides valuable insights into the likely range of illuminant correlated color temperature (CCT) in outdoor scenes.  This information serves to narrow the range of possible illuminant colors. For instance, images taken at noon exhibit a different CCT range than those captured at sunrise or sunset. When combined with additional capture information to distinguish between environments (e.g., indoors vs. outdoors), such metadata can complement image colors to improve illuminant estimation accuracy. 

Despite its potential, a few attempts have explored leveraging additional information available to the camera's image signal processor (ISP) for illuminant estimation, such as metadata-based model control \cite{FFCC} or data augmentation
to enhance generalization \cite{C5}. However, to our knowledge, no previous work has investigated using contextual information---specifically, mobile device timestamps and geolocation---to refine illuminant estimation.

We introduce a method that leverages contextual metadata from mobile phones, along with additional capture information available in camera ISPs, to train a lightweight illuminant estimator model with $\sim$5K parameters. Our model delivers promising results, matching or surpassing that of larger models, while maintaining efficiency. Our model runs on a typical flagship mobile digital signal processor (DSP) and CPU in 0.25 ms and 0.80 ms, respectively. This compact and efficient design is especially advantageous for mobile devices, where minimizing power consumption and memory usage is critical~\cite{FFCC, two-camera, afifi2025optimizing}.

Existing white-balance datasets (e.g., \cite{NUS, gehler2008bayesian, Cube++, IntelTAU}) lack the contextual information needed to validate our method. The absence of contextual information is because most available datasets (e.g., \cite{NUS, Cube++}) used DSLR cameras, which typically lack built-in GPS functionality or accurate timestamps. To address this, we captured a new dataset of 3,224 images using a consumer smartphone camera, accompanied by contextual and capture information. The ground truth for the dataset was established using two approaches: (1) the conventional approach, selecting gray patches from a calibration color chart, and (2) a manually selected user-preference white-balance target, which enhances the image's aesthetic appeal and accounts for incomplete chromatic adaptation \cite{tominaga2014prediction}. The user-preference white-balance ``ground truth'' was validated through a user study. The dataset spans a variety of lighting conditions, including non-standard artificial illuminants \cite{ulucan2022color}, and covers different times of day---sunrise, noon, sunset, and night.

\vspace{5pt}\noindent\textbf{Contribution:}
In this paper, we propose an AWB method that utilizes smartphone contextual metadata (i.e., timestamp and geolocation) alongside capture information to enhance illuminant estimation. We demonstrate that integrating this additional data with conventional color information into a lightweight neural model leads to significant improvements in accuracy. Additionally, we introduce a large-scale dataset captured with a consumer smartphone at various times of day, with ground truth derived from both a calibration color chart and user-preference-based white balance. Benchmarking results show that our method achieves state-of-the-art results in most standard metrics (i.e., angular error statistics \cite{NUS}), while maintaining computational efficiency on mobile devices. 

\section{Method}
\label{sec:method}

\begin{figure}[t]
\centering
\includegraphics[width=\linewidth]{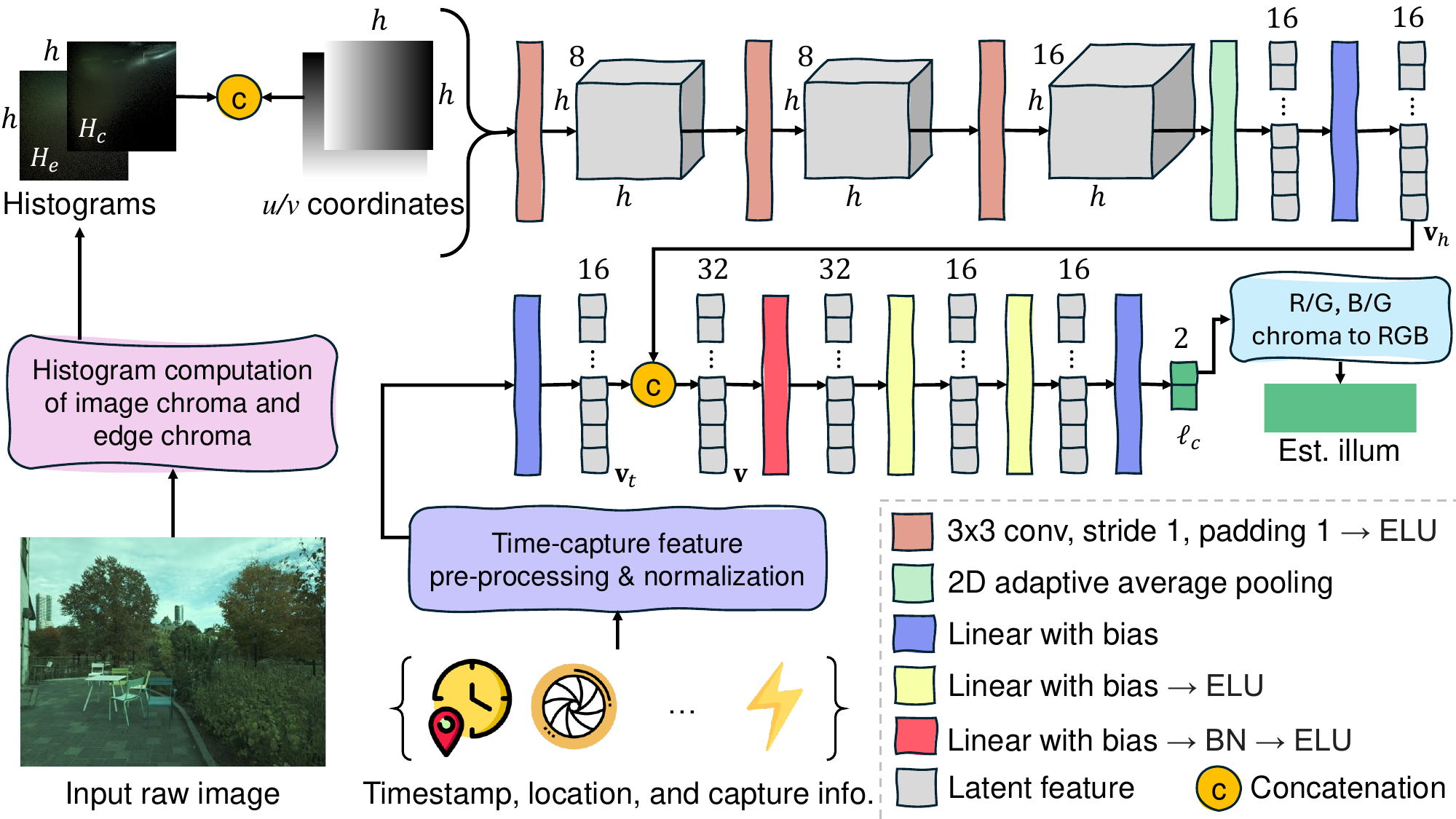}
\vspace{-4mm}
\caption{Our method includes a lightweight model consisting of a convolutional network that processes the histogram feature (derived from raw image colors and edge histograms concatenated with the $u$/$v$ coordinates) to produce a latent feature $\mathbf{v}_{h} \in \mathbb{R}^{16}$. This is then concatenated with the latent feature of the processed time-capture feature, $\mathbf{v}_{t} \in \mathbb{R}^{16}$. The combined feature, $\mathbf{v} \in \mathbb{R}^{32}$, is passed through a lightweight MLP to produce the chromaticity of the scene illuminant, $\mathbf{\ell}_{c} \in \mathbb{R}^{2}$, which is finally converted into normalized RGB illuminant color.\label{fig:main}}
\vspace{-2mm}
\end{figure}

The overview of our method is shown in Fig.~\ref{fig:main}. Our method employs a learnable model that processes two distinct inputs: (1) the time-capture feature (Sec.~\ref{sec:capture-feature}), which combines the contextual and capture information available on mobile devices and accessible by their camera ISPs, and (2) the histogram feature (Sec.~\ref{sec:histogram-feature}), which represents the image's R/G and B/G chromaticity values.
The time-capture feature is first processed to project it into a latent space, producing the time-capture latent feature vector, $\mathbf{v}_{t} \in \mathbb{R}^{16}$. The model processes the histogram feature to produce the histogram's latent feature vector, $\mathbf{v}_{h} \in \mathbb{R}^{16}$. Both the histogram and time-capture latent feature vectors are concatenated and processed by a set of learnable layers to output an R/G, B/G chromaticity vector, $\mathbf{\ell}_{c} \in \mathbb{R}^{2}$, representing the scene illuminant. This 2D vector is then converted into the illuminant RGB color used to perform white balancing.

\subsection{Time-capture feature}
\label{sec:capture-feature}

Our model incorporates contextual metadata, available on mobile devices, as one of its input features. Specifically, we use the geolocation and timestamp of image capture to compute the ``probability'' of the time of day (e.g., sunset, noon, etc.). 
Our approach converts conventional clock time (hour::minutes) to its corresponding {\it time of day} condition based on the date and geolocation of the scene. This approach allows the model to generalize across different time zones and prevents it from being influenced by the capturing location.
The time probability vector represents the likelihood that the captured image corresponds to one of the solar event times (i.e., dawn, sunrise, noon, sunset, dusk, or midnight) and is computed as follows:

\begin{equation}
\label{eq:time-prob1}
p_{g} = 1 - \frac{|t_c - t_g|}{t_s},
\end{equation}

\noindent where $t_c$ is the capture time in seconds (adjusted to the local time zone based on geolocation information), and $t_g$ is the local time in seconds of the solar event, $g \in \{\texttt{dawn}, \texttt{sunrise}, \texttt{noon}, \texttt{sunset}, \texttt{dusk}, \texttt{midnight}\}$, computed using geolocation-based standard algorithms \cite{meeus1991astronomical, walraven1978calculating, van1979low}. The scalar $t_s$ represents the total number of seconds in a day (i.e., 86,400). 

We pre-process the probability of each solar event, $p_{g}$, in our time probability vector by computing the square root to enhance feature representation---compressing high probabilities and amplifying lower ones, which is intended to help create a more balanced and smooth distribution for the model to leverage. We then augment this time probability vector with a one-hot vector, $\mathbf{b}$, that indicates whether the capture time $t_c$ occurs before each solar event. This distinction helps the model account for expected variations in CCT, as illuminant colors can differ before and after certain solar events, such as sunset and sunrise. The value of this one-hot vector for a given solar event $g$ is computed as:
\begin{equation}
\label{eq:time-prob2}
\mathbf{b}_g =
\begin{cases}
1, & \text{if } t_c \leq t_g \\
0, & \text{otherwise},
\end{cases}
\end{equation}

\noindent where $\mathbf{b}_g$ corresponds to the entry in the one-hot vector for the solar event, $g$. Both the time probability vector and the one-hot vector together form our time feature, $\mathbf{p} \in \mathbb{R}^{12}$. 

To enrich our feature set with additional capture information available from the camera ISP and help distinguish the capturing environment (e.g., indoor vs. outdoor), we include the following features in our final time-capture feature, $\mathbf{c}$:
\begin{itemize}
    \item \textbf{ISO} ($i$): The sensitivity of the camera's image sensor to light, where lower values indicate good lighting conditions (e.g., bright scenes) and higher values suggest low-light environments, such as poorly lit indoor scenes.
    \item \textbf{Shutter speed} ($s$): The amount of time the camera's shutter remains open, allowing light to hit the image sensor. It provides an indication of lighting conditions, alongside the ISO value, $i$.
    \item \textbf{Flash status} ($f$): A binary value indicating whether flash light was used during capturing.
    \item \textbf{Noise information (optional)}: Since image denoising is typically applied before or in parallel with illuminant estimation in camera ISPs \cite{liba2019handheld, hasinoff2016burst, delbracio2021mobile}, we also explore the optional use of explicit noise information from the captured scene. More noise typically indicates low-light conditions, which can provide clues about the lighting color range of the scene. Unfortunately, while noise information is accessible within camera ISPs, obtaining accurate noise information for public use is challenging, as the noise profiles in DNG files are not always reliable \cite{zhang2021rethinking}. To address this, we simulate the noise information using two approaches: noise statistics (stats) and/or signal-to-noise ratio (SNR) stats, which are described below.
\end{itemize}

\vspace{5pt}\noindent\textbf{Noise stats ($\mathbf{n}$):} This represents the noise statistics in the captured raw image. We simulate this by denoising each raw image using Adobe Lightroom and computing the noise stats as the mean and standard deviation of each color channel of the absolute difference between the denoised and noisy raw images. This approach provides a simplified method for estimating noise stats, as the denoised images are typically available within the camera ISPs, but difficult to extract from the DNG files.

\vspace{5pt}\noindent \textbf{SNR stats ($\mathbf{r}$):} This alternative approach measures the noise information in the captured image without the need of a denoised reference. The SNR is computed by applying a $15 \times 15$ sliding window over the raw image and calculated as: $10 \log_{10} \left( \frac{\mu}{\sigma + \epsilon} \right)$, where $\mu$  represents the mean RGB of the $15\!\times\!15$ patch, $\sigma$ is the standard deviation, and $\epsilon$ is a small value added for numerical stability. 

Note that while these two approaches---namely, computing noise stats and SNR stats---yield satisfactory results, there is a wide body of research on noise estimation (e.g., \cite{liu2006noise, pyatykh2012image, pimpalkhute2021digital}), which is beyond the scope of this paper.

Our complete time-capture feature is the combination of these inputs and can be expressed as follows: 
\begin{equation}
\label{eq:time_capture_feature}
\mathbf{c} = \left[\mathbf{p}^T, \log\left(i\right), \log\left(s\right), f, \mathbf{w}^T\right]^T,
\end{equation}
\noindent where $\mathbf{w}$ represents optional noise-related features, which can include either noise stats $\mathbf{n}$, SNR stats $\mathbf{r}$, or both. Alternatively, $\mathbf{w}$ can be omitted if noise information is not considered.

Our time-capture feature is first normalized using min-max normalization (with the min and max values computed from the training data), and then processed through a learnable function, $f_{t}$, as follows:

\begin{equation}
\label{eq:time_feature_to_latent}
\mathbf{v}_{t} = f_{t}(\mathbf{c}),
\end{equation}

\noindent where $f_{t}$ is a learnable linear layer that transforms the time-capture feature, $\mathbf{c}$, into its latent representation, $\mathbf{v}_{t} \in \mathbb{R}^{16}$ (see Fig.~\ref{fig:main}).

\subsection{Histogram feature}
\label{sec:histogram-feature}
In addition to the time-capture feature, $\mathbf{c}$, we provide our neural model with a histogram feature, $\mathbf{H}$, which represents the colors of the raw image. Inspired by prior work \cite{CCC, FFCC, C5}, we use a 2D histogram to represent the R/G and B/G chromaticities of the input raw image, $\mathbf{I} \in \mathbb{R}^{K\times3}$, where $K$ denotes the total number of pixels in the raw image. Specifically, we compute the 2D chromaticity histogram, $\mathbf{H}_{c} \in \mathbb{R}^{h\times h}$, as follows:
\begin{equation}
\label{eq:hist1}
    \mathbf{H}_{c}^{(m, n)} = \sum_k \|\mathbf{I}^{(k)}\|_2 \cdot \delta^{(k)}_{m, n},
\end{equation}

\begin{equation}
\label{eq:hist2}
\resizebox{0.9\linewidth}{!}{$%
    \delta^{(k)}_{m, n} = \left[ u_m \leq rg^{(k)} < u_{m+1} \right] \wedge \left[ v_n \leq bg^{(k)} < v_{n+1} \right],
    $}
\end{equation}

\noindent where $rg^{(k)}$ and $bg^{(k)}$ are the R/G and B/G chromaticity values of pixel $k$ in $\mathbf{I}$, and $\|\mathbf{I}^{(k)}\|_2$ represents the intensity of pixel $k$ (i.e., Euclidean norm of the pixel's RGB values). The notation  $\wedge$ denotes the logical AND operator, and $\left[\cdot\right]$ is the Iverson bracket, which evaluates to 1 when the condition is true and 0 otherwise. The histogram bins are defined by the edges $\{u_m\}$ and $\{v_n\}$, where $u_m$ and $v_n$ are the lower edges of the bins, and $u_{m+1}$ and $v_{n+1}$ are the corresponding upper edges. The histogram accumulates the brightness values of pixels whose chromaticity values fall within the range $[u_m, u_{m+1})$ along the R/G axis (i.e., the $u$-axis) and $[v_n, v_{n+1})$ along the B/G axis (i.e., the $v$-axis). $h$ is the number of histogram bins along each axis. Following \cite{CCC}, we compute the square root of the histogram to enhance the utility of the histogram feature \cite{arandjelovic2012three}.

Note that this histogram differs from the $log$-$uv$ histogram used in prior work \cite{CCC, FFCC, C5}, which operates in the logarithmic space of G/R and G/B \cite{finlayson2001color}. We found that the R/G, B/G chromaticity histogram performs better, as it aligns with our model's output space. See the supplemental material for an ablation study. 

In addition to the chromaticity histogram, $\mathbf{H}_{c}$, of the image colors, and building on prior work \cite{FFCC, C5}, we augment it with the square-rooted chromaticity histogram of the image's edges, $\mathbf{H}_{e}$, where the image edges are computed as follows:

\begin{equation}
\label{eq:edge}
\mathbf{E}^{(x, y)} = \frac{1}{8} \sum_{\Delta x, \Delta y} \left| \mathbf{I}'^{(x, y)} - \mathbf{I}'^{(x + \Delta x, y + \Delta y)} \right|,
\end{equation}

\noindent where $\mathbf{E}$ represents the image edges, $\mathbf{I}'$ refers to the raw image in 3D tensor form (height, width, channels), and $\Delta x, \Delta y \in \{-1, 0, 1\}$ with $(\Delta x, \Delta y) \neq (0, 0)$. 
The edge histogram, $\mathbf{H}_{e}$, is computed from $\mathbf{E}$ using Eqs.~\ref{eq:hist1} and \ref{eq:hist2}.

Our histogram feature is constructed by concatenating these two histograms, $\mathbf{H}_{c}$ and $\mathbf{H}_{e}$. Since this histogram feature is first processed by convolutional (conv) layers, as shown in Fig.~\ref{fig:main}, we follow \cite{C5} by appending additional channels that encode the positional information of the $u$/$v$ coordinates in histogram space, which helps capture spatial relationships within the histogram feature. Consequently, our final histogram feature, $\mathbf{H} \in \mathbb{R}^{h\times h\times4}$, consists of 2 channels representing the chromaticity of the image and its edges, along with the additional  $u$/$v$ coordinate channels. This feature is processed through a series of conv layers with ELU activation \cite{clevert2015fast}. The resulting latent representation undergoes adaptive average pooling before passing through a linear layer to produce the histogram's latent feature vector, $\mathbf{v}_{h} \in \mathbb{R}^{16}$, as follows:

\begin{equation} \label{eq:hist_feature_to_latent} \mathbf{v}_{h} = f_{h}(\mathbf{H}), \end{equation}

\noindent where $f_{h}$ denotes the sub-model (i.e., conv layers, ELU activation, pooling, and linear layer) that maps the histogram feature into its latent space, as shown in Fig.~\ref{fig:main}.

\subsection{Illuminant estimation}
\label{sec:optmization}

Both feature vectors, $\mathbf{v}_{t}$ and $\mathbf{v}_{h}$, are concatenated to produce the latent vector $\mathbf{v}$ and processed by an illuminant estimation sub-model as follows: 

\begin{equation} \label{eq:cat} \mathbf{v} = \left[\mathbf{v}_t;\mathbf{v}_h\right], \end{equation}

\begin{equation} \label{eq:cat} \mathbf{\ell}_c = f_{\ell}(\mathbf{v}), \end{equation}

\noindent where $f_{\ell}$ consists of a set of linear layers, with batch normalization applied to the first layer, followed by activation functions---except for the final layer, which outputs $\mathbf{\ell}_c$, the chromaticity vector of the scene illuminant. This vector is then transformed into an unnormalized RGB illuminant color by mapping $[$R/G, B/G$]^T \to [$R/G, 1, B/G$]^T$, followed by normalization via division by its L2 norm to produce the final illuminant color. We optimize $f_t$, $f_h$, and $f_{\ell}$ to minimize the angular error \cite{hordley2004re} between the predicted RGB illuminant color and the ground-truth RGB illuminant color.

\section{Dataset}
\label{sec:dataset}

\begin{figure}[!t]
\centering
\includegraphics[width=\linewidth]{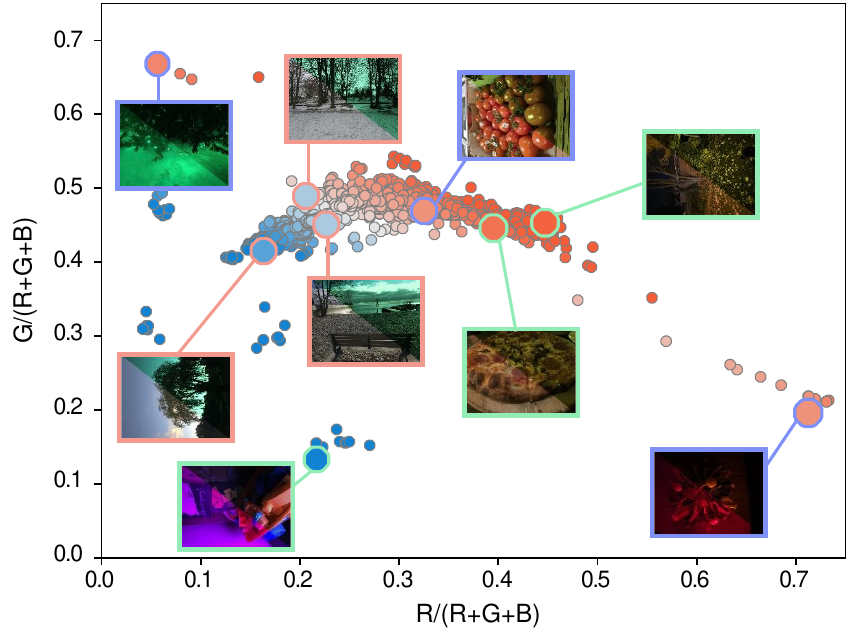}
\vspace{-5mm}
\caption{$rg$ chromaticity distribution of ground-truth illuminant colors for neutral white balance in our dataset. We show example images in raw and sRGB spaces, with raw images gamma-corrected for better visualization. See the supplemental material for the $rg$ chromaticity distribution of user-preference illuminants. \label{fig:dataset1}}
\vspace{-3mm}
\end{figure}

To train and validate our method, we require a dataset that includes contextual information (i.e., timestamp and geolocation) for each image. Existing datasets (e.g., \cite{NUS, gehler2008bayesian, Cube++, IntelTAU, NCC, li2024nightcc}) lack this essential information, motivating us to collect a new dataset using a smartphone camera that provides contextual metadata for each image. Specifically, we captured 3,224 linear raw images with the Samsung S24 Ultra's main camera, covering a wide range of scenes both indoors and outdoors, at various times of day (e.g., sunset, sunrise, noon, night). Our dataset includes images captured under various light sources (e.g., sunlight, incandescent, LED), as well as non-standard illuminant colors (e.g., colored LED light), which are not present in existing datasets \cite{ulucan2022color}; see Fig.~\ref{fig:dataset1}. Additionally, our dataset captures scenes under different weather conditions (sunny, cloudy, rain, snow, etc.). Example scenes can be found in the supplemental material.

\begin{figure*}[t]
\centering
\includegraphics[width=\linewidth]{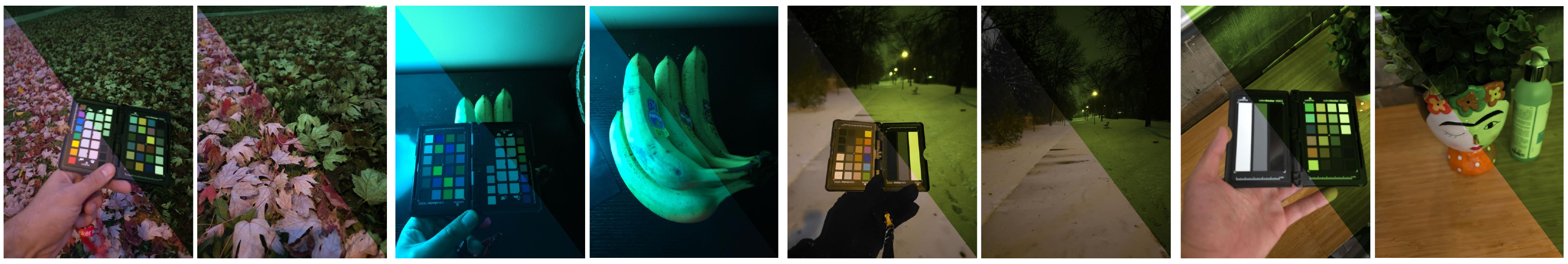}
\vspace{-5mm}
\caption{For each scene in our dataset, an image with a color chart placed in the scene was captured. The gray patches on the color chart were used to measure the ground-truth illuminant. These color-chart images were discarded, and only the images of the scenes without the color chart were used for the training, validation, and testing sets. For each example, we show both raw and sRGB images, with the raw images gamma-corrected to enhance visualization.\label{fig:color-chart}}
\vspace{-2mm}
\end{figure*}

We follow prior work \cite{IntelTAU, afifi2025optimizing} in collecting ground-truth illuminant colors for each scene. Specifically, for each scene, we first capture an image with a calibration color chart, which is used to extract the illuminant color from the gray patches. Next, we capture an image of the same scene without the color chart (see Fig.~\ref{fig:color-chart}), resulting in $\sim$6K images. After obtaining the ground-truth illuminant color, we discard all color chart images, leaving us with 3,224 images in our dataset. This approach allows us to test with natural images that mimic real-world scenarios, without the need to mask out the color chart patch.

Since our dataset includes a wide variety of lighting conditions, such as sunset/sunrise, night scenes, and artificial light, we generate a ``user-preference'' ground truth in addition to the neutral ground truth obtained from the color chart for each scene in our dataset.  This is to account for human incomplete chromatic adaptation in such scenes \cite{incomplete_cat1, tominaga2014prediction} and user preference \cite{ershov2022ntire, choi2014user}. Specifically, an expert photographer was asked to assign a ground truth illuminant to each scene to make it appear more natural, reflect real-world observations, and enhance the aesthetics of the image. Notably, the same person who captured the scene also performed the annotation, ensuring that the user-preference selection was based on their real-world observations of how the scene should appear. 

We validated the user-preference ground truth through a study with 20 participants, who selected the most natural image from pairs of white-balanced images corresponding to the user-preference ground truth and the neutral ground truth (obtained from the color chart). The user-preference ground truth was selected in 71.95\% of the trials, confirming its preference. See the supplemental material for additional details. In our experiments, both ground-truth types---neutral and user-preference---were used for training and evaluation (Sec.~\ref{sec:results}).

Additionally, we created a mask for regions illuminated by non-dominant illuminants in each scene. These masks ensure that all scenes have only one dominant illuminant, matching the color of the neutral ground truth, without confounding effects from other illuminants. We applied these masks to the training images when training our method and others. To preserve privacy, sensitive information, such as car plates and faces, has been blurred.

In addition to raw images, both ground-truth illuminant types, and contextual and capture information, we provide additional valuable auxiliary data to broaden the impact of our dataset. This includes locally tone-mapped sRGB images rendered by an expert, that can serve as ground-truth for applications beyond white-balance correction,
such as neural ISPs \cite{xing2021invertible, ignatov2022microisp, kim2023paramisp, he2024enhancing}. More information about the dataset can be found in the supplemental material. We organize our dataset into 2,619 raw images for training, 205 raw images for validation, and 400 raw images for testing. 

\section{Experiments}
\label{sec:results}

\input{tables/results_test_wo_mask_w_params}

\vspace{5pt}\noindent \textbf{Implementation details:}
We train the model using the Adam optimizer \cite{adam}, for 400 epochs with betas set to (0.9, 0.999) and a weight decay factor of $10^{-9}$. A warm-up strategy is applied to gradually increase the learning rate from $10^{-6}$ to $10^{-3}$ over the first 5 epochs. After this, we use a cosine annealing schedule \cite{cosineAnnealing}. Following \cite{C5}, we employ a batch-size increment strategy during training, starting with a batch size of 8 and doubling it every 100 epochs. 
Following prior work \cite{C5, afifi2025optimizing}, we use images of size 384$\times$256 pixels in all experiments, which is a reasonable size for camera pipelines to reduce computational overhead. For our method, we use histograms with $h=48$ bins. The histogram boundaries are determined by computing the 10th and 95th percentiles of the chromaticity values along each chromaticity axis from the training set. We report the results of our method both with and without the inclusion of noise stats, $\mathbf{n}$, and SNR stats, $\mathbf{r}$.

\vspace{5pt}\noindent\textbf{Results:} We report the results of our method alongside several others benchmarked on the proposed dataset. Our method is compared with various statistical-based approaches \cite{GW, SoG, GE, maxRGB, wGE, NUS, MSGP, GI, TECC, RGP}, camera-specific learning-based methods \cite{GAMUT, NIS, CLASSIFICATION, FFCC, FC4, APAP, knn, BoCF, C4, CWCC, TLCC, PCC, CFCC}, and camera-independent techniques \cite{QUASI-CC, SIIE, C5}. For the camera-specific learning-based approaches, including our method, all models were trained on our training set.

For the camera-independent methods \cite{SIIE, C5}, we present results from models trained on the NUS \cite{NUS} and Cube++ \cite{Cube++} datasets. Additionally, we report results from fine-tuned versions of these models \cite{SIIE, C5}, incorporating our proposed dataset. We also evaluate camera-specific (CS) models of these methods \cite{SIIE, C5}, which were trained exclusively on our training data, without any additional datasets. For further details, refer to the supplemental material.

Table~\ref{tab:results-test-wo-mask} shows the results on the testing set of our dataset. We report the mean, median, best 25\%, worst 25\%, worst 5\%, tri-mean, and maximum angular errors between the estimated illuminant colors and the ground truth colors for each method. Results are provided for both neutral and user-preference ground truth, with two models, one trained for each ground-truth type, except for the models SIIE \cite{SIIE}, SIIE (tuned) \cite{SIIE}, C5 \cite{C5}, and C5 (tuned) \cite{C5}, which were trained on images from the NUS \cite{NUS} and Cube++ \cite{Cube++} datasets that do not include user-preference ground truth.

Additionally, we report the total number of parameters for methods that involve tunable or learnable parameters. The results in Table~\ref{tab:results-test-wo-mask} are based on images without masking regions illuminated by a light source other than the dominant one used to obtain the ground truth, reflecting real-world scenarios where scenes with a single light source are not always guaranteed. For completeness, additional results are provided in the supplementary material, where regions illuminated by light sources different from the ground-truth illuminant color are masked out.

To further demonstrate the effectiveness of our method on existing DSLR datasets, where contextual metadata is typically unavailable, we report our results on the Simple Cube++ dataset \cite{Cube++}. Although geolocation metadata is absent, the dataset provides capture settings such as ISO and exposure time from the DSLR camera ISP. 

We trained our model using the histogram feature $\mathbf{H}$, along with capture features including ISO ($i$), exposure time ($e$), and SNR stats ($\mathbf{r}$), deliberately excluding the contextual metadata that is not available in the Simple Cube++ DSLR dataset. The results of our method, both with and without the SNR stats, are compared to other methods in Table~\ref{tab:results-cube}. As shown, our method outperforms competing methods across most evaluation metrics.

\input{tables/results_cube}

\vspace{5pt}\noindent\textbf{Inference time:} As shown in Table~\ref{tab:results-test-wo-mask}, our method performs comparably to or outperforms prior methods, while maintaining a compact model with only approximately 5K parameters. This lightweight design results in faster runtimes compared to other methods that achieve competitive results, namely C4 \cite{C4} and tuned C5 \cite{C5}. 

In Table~\ref{tab:processing-time}, we present the processing times of our model, C4, and C5 on an AMD Ryzen Threadripper PRO 3975WX CPU and an NVIDIA RTX A6000 GPU. Our fast runtime performance is particularly well-suited for mobile camera ISPs, where limited computational latency is crucial due to the processing demands of other modules (e.g., denoising, local tone mapping) per frame. Our model runs in just 0.25 ms on the DSP and 0.80 ms on the CPU of the Samsung S24 Ultra.

\input{tables/processing_time}

\vspace{5pt}\noindent \textbf{Ablation studies:}
We conducted ablation studies to evaluate the impact of each input feature on the validation set of our dataset. Specifically, we assessed our method by excluding either the time-capture feature ($\mathbf{c}$) or the histogram feature ($\mathbf{H}$). Additionally, we tested the method using only the time feature ($\mathbf{p}$) and noise stats ($\mathbf{n}$), while excluding the histogram feature ($\mathbf{H}$). We also evaluated the method using the histogram feature along with the time feature and noise stats, excluding other capture information. Furthermore, we examined the accuracy of our method using all input features except for $\mathbf{p}$. Lastly, we present the results when all input features are used, but without the noise stats ($\mathbf{n}$) and the SNR stats ($\mathbf{r}$). These results are presented in Table~\ref{tab:results-validation-ablation}.

As shown in Table~\ref{tab:results-validation-ablation}, using only the time feature ($\mathbf{p}$) and noise stats ($\mathbf{n}$) yields a reasonable accuracy (2.37$^\circ$ mean angular error), compared to FFCC \cite{FFCC}, which achieves 2.19$^\circ$ mean angular error on the validation set. This validates the usefulness of time-of-day and noise information in providing contextual clues. As expected, incorporating the histogram feature ($\mathbf{H}$), which represents scene colors, along with the time-of-day and noise stats, significantly boosts accuracy, as demonstrated by the results in the fourth row. We further investigate the impact of using noise information in the last three rows: using SNR stats ($\mathbf{r}$), using noise stats ($\mathbf{n}$), or both. The combination of both noise features yields the lowest mean angular error. Additional ablation studies are provided in the supplementary material.

\input{tables/val_ablation}

\section{Conclusion and limitations}
\label{sec:conclusion}
We presented a method for in-camera AWB correction that leverages contextual information. Specifically, we proposed a lightweight model that leverages contextual metadata (notably time-of-day derived from timestamp and geolocation) to guide the illuminant estimation process. In addition to this contextual metadata, we incorporate image colors in the form of histogram features, as well as capture information such as ISO, shutter speed, and noise stats, to help the model distinguish between artificial and natural scenes and improve its final accuracy. Our method is fast and can achieve high frame rates on modern smartphone DSPs and CPUs, while maintaining accurate illuminant color estimation. 

A key contribution of this work is a large-scale dataset of raw images captured by a consumer smartphone camera, along with the necessary contextual metadata for training and evaluating our method. Beyond the traditional neutral white-balance ground truth extracted from a calibration color chart placed in each scene, we also include a user-preference ground truth that targets the observer's preference, validated through a user study. Results based on both ground truth types demonstrate that our method achieves comparable or superior performance to existing methods, which require larger models.

While our method represents a promising solution for mobile camera ISPs, its dependency on contextual metadata limits its optimal accuracy to devices that provide geolocation data, which may not available on most DSLR cameras. Additionally, while the contextual metadata is device-independent (i.e., the differences across devices are expected to be minimal), our method relies on additional capture information (i.e., ISO, shutter speed, noise, and SNR stats) and image colors, making it inherently camera-dependent in design. This dependency prevents our trained model from generalizing to new devices without fine-tuning or re-training. However, this issue can be mitigated through color calibration (e.g., \cite{lin2023color}), which can be extended to calibrate capture information as well---by performing a pre-processing mapping from the new camera space (for both color and capture information) to the camera space used during training. Additional discussion on cross-camera generalization is provided in the supplementary material.


\clearpage
\maketitlesupplementary
\appendix

\begin{figure*}[!t]
\centering
\includegraphics[width=\linewidth]{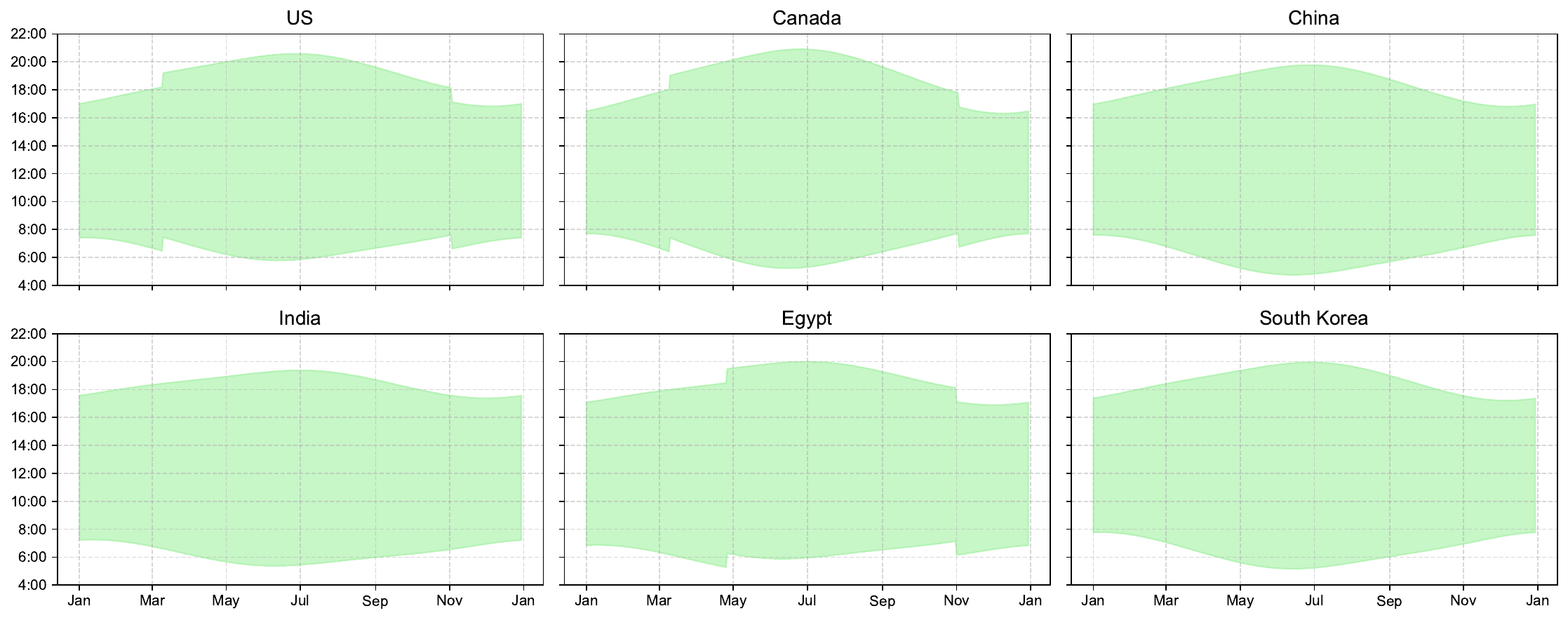}
\vspace{-4mm}
\caption{This figure illustrates the daily variations in sunrise and sunset times across different countries throughout the year 2024. The x-axis represents the date, while the y-axis denotes the time of day (4 AM – 10 PM). The light green shaded regions indicate the duration of daylight for each country, highlighting seasonal variations due to differences in latitude and geographical location. \label{fig:sunset_location}}
\end{figure*}

\begin{figure*}[!t]
\centering
\includegraphics[width=\linewidth]{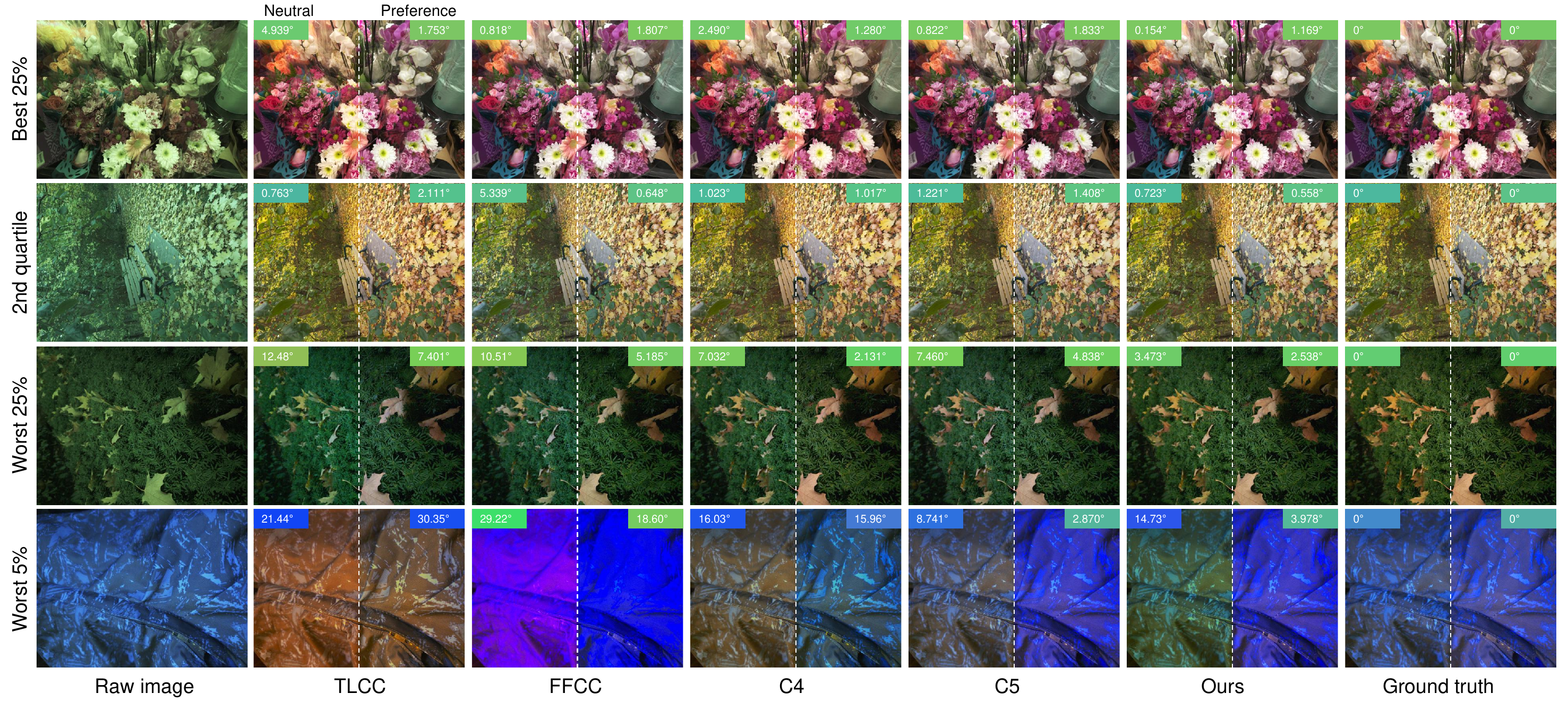}
\vspace{-4mm}
\caption{Qualitative examples from the best 25\% (first quartile), second quartile, third quartile, and the worst 5\% of our results. Shown images are white-balanced using the illuminant estimates from TLCC \cite{TLCC}, FFCC \cite{FFCC}, C4 \cite{C4}, C5 \cite{C5}, and our method. We show results of both types of white-balance corrections: 1) neutral (on the left side of each white-balanced image) and 2) user-preference (on the right side). All images are gamma-corrected to enhance visualization.\label{fig:qualitative}}
\end{figure*}

This supplementary material provides additional details on the experiments presented in the main paper (Sec.~\ref{supp-sec:comparisons}), additional details on the contextual information, ablation studies, and results (Sec.~\ref{supp-sec:results}), and detailed information about our dataset (Sec.~\ref{supp-sec:dataset}). Lastly, we provide additional ground-truth data to broaden the dataset’s impact for other applications (Sec.~\ref{sec:additional-gt-data}).

\begin{figure*}[t]
\centering
\includegraphics[width=\linewidth]{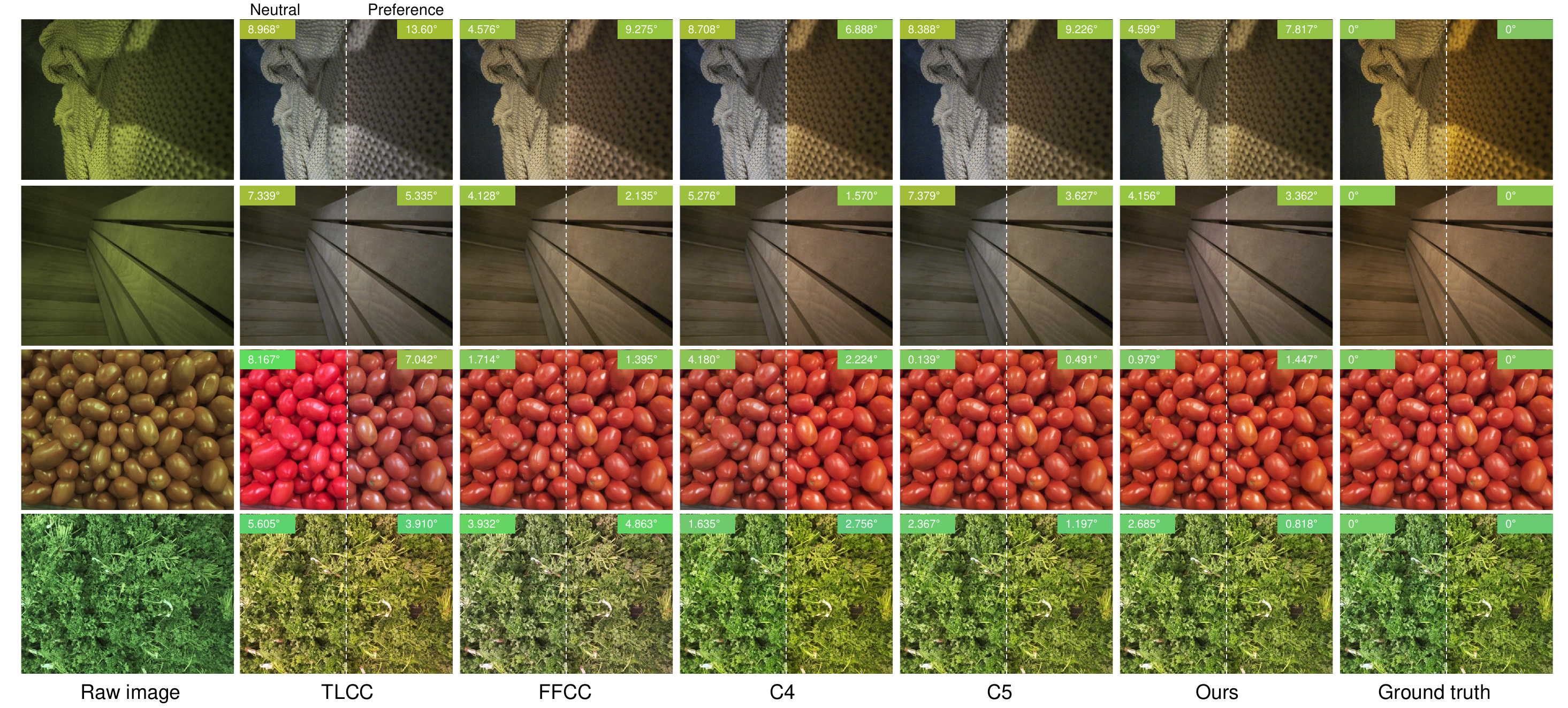}
\vspace{-4mm}
\caption{Additional qualitative examples of scenes with limited color variations, which are particularly challenging for illuminant estimation. Images are white-balanced using illuminant estimates from TLCC \cite{TLCC}, FFCC \cite{FFCC}, C4 \cite{C4}, C5 \cite{C5}, and our method. For reference, we also include results corrected using the ground-truth illuminant. We show results of both types of white-balance corrections: 1) neutral (on the left side of each white-balanced image) and 2) user-preference (on the right side). All images are gamma-corrected for better visualization. \label{fig:qualitative-dominant-color}}
\end{figure*}

\section{Additional comparison details}\label{supp-sec:comparisons}

In the main paper and this supplementary material, we report the results of various methods evaluated on our proposed dataset. We benchmark several approaches, including statistical-based (learning-free) methods, camera-specific learning-based methods, and cross-camera learning-based methods. 

For the cross-camera learning-based methods (specifically, SIIE \cite{SIIE} and C5 \cite{C5}), we report results for three versions of each method:

\begin{itemize} 
\item A model trained on the Cube++ \cite{Cube++} and NUS \cite{NUS} datasets (results reported without any postfix). Since the Cube++ and NUS datasets lack user-preference ground truth, we only report results on neutral ground truth using our dataset. 
\item A model trained on the Cube++ and NUS datasets, as well as our proposed dataset (results reported with the postfix (tuned)). Similarly, due to the absence of user-preference ground truth in the Cube++ and NUS datasets, we only report results on neutral ground truth using our dataset.
\item A model trained exclusively on our dataset (results reported with the postfix (tuned-CS), where 'CS' stands for camera-specific).
\end{itemize}

This approach ensures a fair comparison, as the generic models (trained on Cube++ and NUS datasets) lack exposure to the diverse lighting conditions present in our dataset.

For the C5 method \cite{C5}, when training the tuned-CS model, we used only the input histogram and excluded the additional histograms proposed in the original method. These additional histograms were mainly intended to assist the model in calibrating for new cameras. Since our dataset uses a single camera, we removed the extra encoders from the C5 model for the tuned-CS version.

For FFCC \cite{FFCC}, we first performed tuning to identify optimal hyperparameters before training the model. The model was tuned and trained on our dataset. When reporting FFCC with capture metadata---denoted as FFCC (capture info) in the tables---we used a vector of 
$[log(\text{shutter-speed}), log(\text{ISO}), 1]$ instead of the original metadata vector $[log(\text{shutter-speed}), log(\text{f-number}), 1] \times [\text{cam-1}, \text{cam-2}, 1]$, for the following reasons: we only have a single camera (the original method was designed to handle two different cameras), and our dataset focuses on smartphone cameras, which have a fixed aperture (no change in f-number per scene).

For the classification-CC method \cite{CLASSIFICATION}, we re-implemented the approach as the original code was unavailable. In our implementation, we set the number of clusters to 50, matching the value used in the original paper \cite{CLASSIFICATION} for the NUS dataset \cite{NUS}.

For the KNN method \cite{knn}, which was initially proposed for white-balance correction in the post-capture stage, we followed the adjustments used in the evaluation presented in \cite{SIIE}. Specifically, we replaced the polynomial function used in the original method with the ground-truth 3D illuminant vectors from the training data. The nearest-neighbor process was performed as described in the original paper, but the final output was an illuminant color, rather than a polynomial function.

For the quasi-unsupervised CC method \cite{QUASI-CC}, we report the results of both the unsupervised model and the tuned model on our dataset. We used the gray-world (GW) method \cite{GW} as the initial estimation for APAP \cite{APAP}. 

For the TLCC method \cite{TLCC}, we used the official checkpoint released by the authors, trained on the sRGB dataset \cite{Places205} and raw datasets \cite{gehler2008bayesian, NUS}. We then finetuned the model on our proposed dataset to leverage transfer learning from sRGB to raw, as described in the TLCC paper \cite{TLCC}.

For the gamut method \cite{GAMUT}, we present the results for three canonical gamuts: edges, pixel colors, and the 1st-degree gradient. For the TECC method \cite{TECC}, we report the results with the 2nd-order gray-edge \cite{GE}. Lastly, for the gray-edge (GE) method \cite{GE}, we provide the results based on both the first and second gradients of the images.

\begin{figure*}[!t]
\centering
\includegraphics[width=\linewidth]{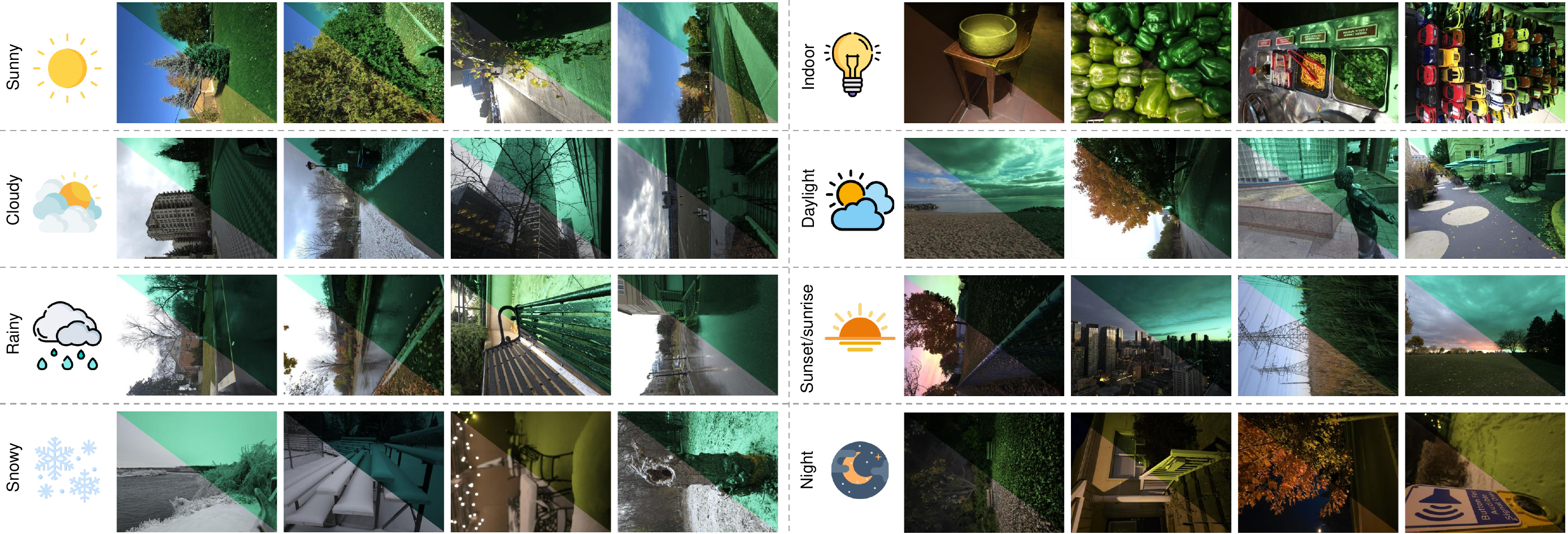}
\vspace{-4mm}
\caption{Our dataset includes diverse scenes captured under various weather conditions (sunny, cloudy, rainy, and snowy) and lighting conditions (indoor, daylight, sunset/sunrise, and night). For each example, we show raw images (gamma-corrected for better visualization) alongside their sRGB counterparts. \label{fig:dataset_examples}}
\end{figure*}

\section{Additional details and results}
\label{supp-sec:results}

\subsection{Additional details on contextual information}

In the main paper, we presented our method, which relies on the ``probability'' of an image being captured during one of the key solar events (e.g., sunrise, noon, sunset) along with additional capture metadata and color information represented as histograms. 

Our method leverages the probability of the time of day, allowing the model to rely on an absolute time reference rather than being affected by location-specific time zones, thereby improving generalization. 

Solar event times (i.e., dawn, sunrise, noon, sunset, dusk, and midnight) vary significantly based on the time of year and geographical location. For instance, locations near the equator experience relatively small variation in day length, whereas higher-latitude regions exhibit more pronounced seasonal differences. In Fig.~\ref{fig:sunset_location}, we illustrate the average length of daylight across different countries and continents. As is well known, sunset/sunrise times vary depending on both location and date.

If we were to use the raw clock timestamp without geolocation, the information would be highly location-dependent and would not generalize well to regions with different solar event timings. An alternative approach would be to provide both geolocation and timestamp, allowing the model to learn their relationship with solar event timings. However, this would require a diverse dataset with images captured across different locations worldwide to ensure robust learning, which may be impractical due to the extensive data collection required. Our method is simpler and more effective--instead of relying on learned patterns, we use traditional astronomical methods \cite{meeus1991astronomical, walraven1978calculating, van1979low} to compute solar event times for a given location. This allows us to represent time in an absolute manner, using the probability of an image being captured at each solar event rather than relying on location-specific timestamps.

\input{tables/results_val_w_mask}

\subsection{Additional ablation studies}

In the main paper, we presented a set of ablation studies to analyze the impact of different input features on our method. Here, we provide additional ablation studies on the validation set with masks applied, as shown in Table~\ref{tab:results-validation}. By default, our results in Table~\ref{tab:results-validation} use the histogram feature, $\mathbf{H}$, and the time-capture feature, $\mathbf{c}$, with noise stats, $\mathbf{n}$.

In this additional set of ablation studies, we show results when using only the time feature, $\mathbf{p}$, without the histogram feature (w/o $\mathbf{H}$). We then added capture information, including ISO ($i$), shutter speed ($s$), flash status ($f$), and noise stats ($\mathbf{n}$), one at a time, in addition to the time feature, $\mathbf{p}$. Additionally, we examine the effect of using the histogram feature in combination with only ISO ($i$), shutter speed ($s$), flash status ($f$), and noise stats ($\mathbf{n}$). 

Furthermore, we present results using the complete time feature with noise stats, $\mathbf{n}$, under the following conditions:
\begin{itemize} 
\item Without the edge histogram, $\mathbf{H}_{e}$.
\item Without the additional $u$ and $v$ positional encoding channels ($u$/$v$ coord.).
\item Using the $log$-$uv$ histogram from prior work \cite{CCC, FFCC, C5} instead of our R/G and B/G chromaticity histogram.
\item Using the R/G, B/G chromaticity image, $\mathbf{I}_{\texttt{chroma}}$, instead of our histogram feature.
\item Without pre-processing and normalization of the time-capture feature.
\item Using a smaller histogram feature with 24 bins.
\item Using a lower-resolution image of $64\times48$.
\item Without the time feature, $\mathbf{p}$. 
\item Various combinations of noise stats, $\mathbf{n}$, and SNR stats, $\mathbf{r}$.
\end{itemize}

\subsection{Analysis on outdoor vs. indoor scenes} 

For outdoor scenes, time-of-day information provides strong cues about the likely range of illuminants. However, it is not as informative for indoor scenes, which are typically illuminated by artificial lights. Therefore, for indoor scenes, image colors and additional capture information are necessary for accurate illuminant estimation.

In Table~\ref{tab:results-outdoor-indoor}, we analyze the angular error of our model when trained using 1) the time feature $\mathbf{p}$ only, 2) the complete time-capture feature without the color histogram, and 3) the complete time-capture feature with the color histogram, across different scene types. The models are trained on all scene types and tested separately on outdoor and indoor scenes, with the indoor and outdoor scenes manually labeled (see Sec.~\ref{supp-sec:gui-tool} for details).

The first row of Table~\ref{tab:results-outdoor-indoor} presents results for using only the time feature $\mathbf{p}$, without any color information from the scene provided by the histogram $\mathbf{H}$. The results for outdoor scenes are significantly better than those for indoor scenes, indicating that the time feature $\mathbf{p}$ provides valuable cues for estimating illuminants in outdoor scenes.

The second row shows results for using the complete time-capture feature, without scene color information. The outdoor/indoor gap is smaller, suggesting that other capture data, such as ISO and noise stats, provide additional insights into the capturing environment. The third row shows results for using the complete time-capture feature along with the image histogram (our proposed method). This further reduces the outdoor/indoor gap, as the histogram provides color information for both indoor and outdoor scenes.

\input{tables/results_outdoor_indoor_test_wo_mask}

\input{tables/results_test_w_mask}

\subsection{Additional quantitative results}

In the main paper, we reported results on our testing set without masking out regions illuminated by light sources different from the dominant one used to obtain the ground truth. This setup mimics realistic scenarios where a single illuminant is not always present. In Table \ref{tab:results-test}, we report results after masking out regions in the testing set that are illuminated by different light sources than the ground truth. Table \ref{tab:results-validation-wo-mask} shows comparisons with other methods on the validation set without masking out regions lit by different illuminations than the dominant light color in the scene.

\input{tables/results_val_wo_mask}

\subsection{Qualitative results} 

Figure~\ref{fig:qualitative} presents qualitative results from our method alongside other illuminant estimation methods, namely TLCC \cite{TLCC}, FFCC \cite{FFCC}, C4 \cite{C4}, and C5 \cite{C5}. We include randomly selected examples representing the top 25\%, second quartile, third quartile, and bottom 5\% of our results. For FFCC \cite{FFCC} and C5 \cite{C5}, we show the best result from each method for every example shown in the figure, as we used multiple models for each method---FFCC includes models with and without capture information, and C5 has differently tuned models. 

As shown in Fig.~\ref{fig:qualitative}, the worst 5\% example is a scene with limited colors, a typical challenge in illuminant estimation, where color information can mislead any model from achieving accurate estimates. Although our method has a relatively high error, other methods, such as TLCC \cite{TLCC} and FFCC \cite{FFCC}, exhibit even higher errors. However, our method results in more perceptually acceptable differences compared to these methods when compared to the ground truth. 

To further examine our method on scenes with limited colors, we present additional qualitative examples in Fig.~\ref{fig:qualitative-dominant-color}. As shown, our method performs reasonably well in these challenging cases when compared to other methods (e.g., TLCC \cite{TLCC}).


\subsection{Cross-camera generalization} 

In the main paper, we explained that our method is inherently camera-specific by design. However, this limitation can be mitigated through calibration. Here, we present an experimental evaluation of a calibration-based solution to address the camera-specific nature of our approach. Specifically, we calibrate a polynomial mapping for metadata (ISO, shutter speed) and a $3\!\times\!3$ color mapping matrix between our primary camera (Samsung S24 Ultra main camera) and the Samsung S25 Ultra telephoto camera. We evaluate two strategies on 257 test scenes captured by the S25 Ultra telephoto camera:
\begin{enumerate}
    \item Mapping the S24 Ultra main camera's training data to the S25 Ultra telephoto camera's space \textit{offline}, followed by training on the mapped data.
    \item Mapping S25 Ultra telephoto images and metadata \textit{online} to the S24 Ultra main camera’s space, applying the model trained on the S24 main camera, and then mapping the predicted illuminant back to the S25 Ultra telephoto camera’s space.
\end{enumerate}

Results are shown in Table~\ref{tab:cross-camera}, alongside C4~\cite{C4} and C5~\cite{C5}, both of which are cross-camera methods. None of the methods, including ours, had access to training examples from the target camera (S25 Ultra telephoto).

\input{tables/cross_cam}


\section{Additional details of dataset}\label{supp-sec:dataset}

\begin{figure*}[t]
\centering
\includegraphics[width=\linewidth]{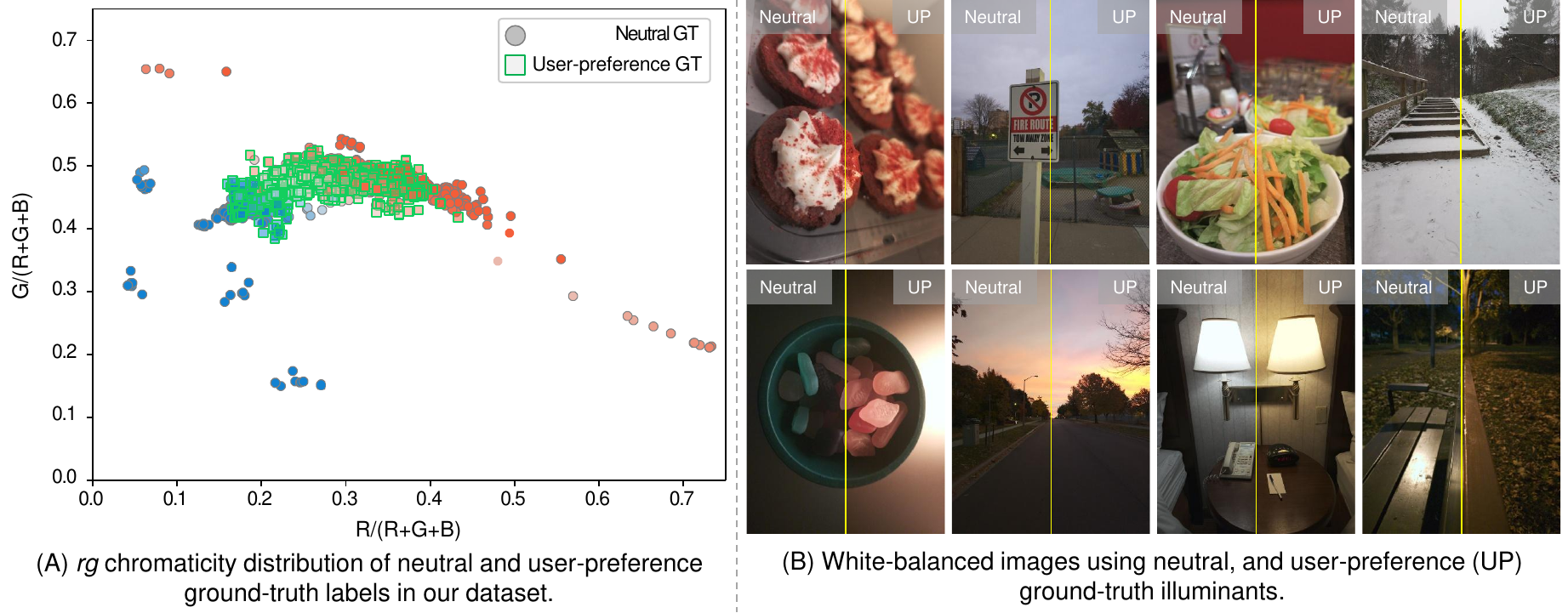}
\vspace{-4mm}
\caption{Our dataset includes the ground-truth illuminant color for each scene under neutral white balance (obtained from gray patches of a color chart) and user-preference white balance, where an expert photographer adjusts the white-balance illuminant color of each image to match real scene observations and enhance image aesthetics. In (A), we show the $rg$ chromaticity distribution of both neutral and user-preference ground-truth illuminants, and in (B), example linear images from our dataset corrected using these ground-truth illuminants. Color correction matrix (CCM) and gamma correction are applied to enhance visualization.\label{fig:user_preference}}
\end{figure*}

\begin{figure}[!t]
\centering
\includegraphics[width=\linewidth]{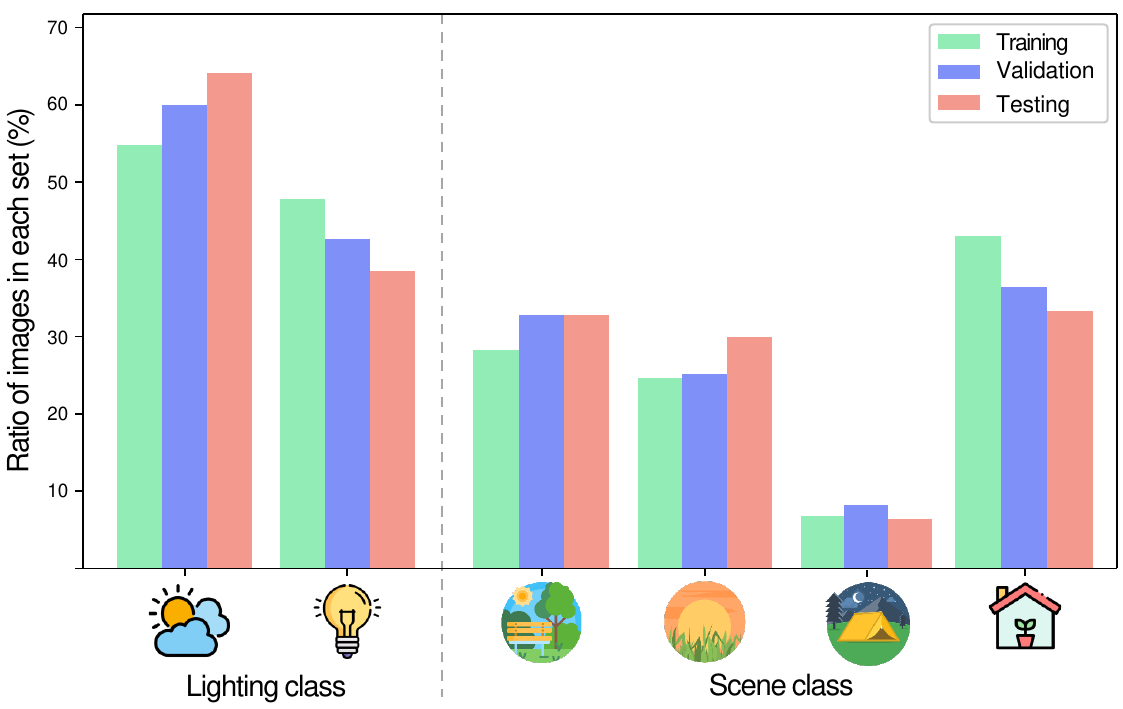}
\vspace{-4mm}
\caption{Statistics of our dataset categorized by lighting classes (`natural' and `artificial' light sources) and semantic scene classes (`outdoor [daylight]', `outdoor [sunset/sunrise]', `outdoor [night]', and `indoor'). \label{fig:categories}}
\end{figure}

\begin{figure*}[!t]
\centering
\includegraphics[width=\linewidth]{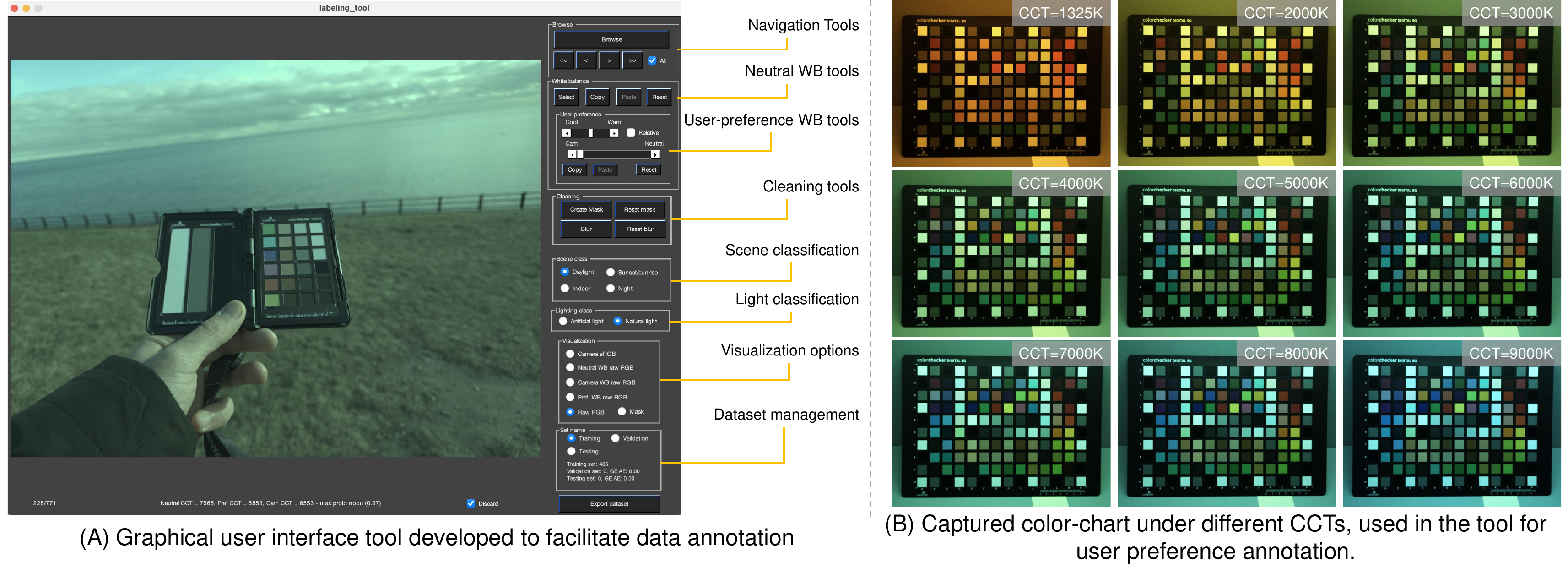}
\vspace{-4mm}
\caption{(A) The graphical user interface (GUI) tool developed to facilitate the annotation process. The tool provides features such as navigation tools, neutral white balance (WB) tools, user preference WB tools, cleaning tools, scene and light classification options, visualization options, and dataset management functionalities. In the neutral WB tools, the annotator can select a reference white point from a raw color chart image, copy it, and paste it into the sequential scene(s) sharing the same lighting condition. In the user preference WB tools, the annotator can interpolate between the camera WB and the neutral WB. Additionally, the annotator can adjust the corresponding correlated color temperature (CCT) of the selected WB setting to create a cooler or warmer appearance. (B) Color charts captured under different CCTs, used within the GUI tool, to calculate the corresponding CCT of illuminant colors in the camera raw space. The images shown in (B) are in raw space with a gamma correction applied to enhance visualization. \label{fig:gui}}
\end{figure*}

In the main paper, we presented our dataset of 3,224 images captured by the Samsung S24 Ultra’s main camera. Example scenes from our dataset are shown in Fig.~\ref{fig:dataset_examples}. As shown, our dataset includes diverse scenes captured under various weather and lighting conditions. 

A distinctive feature of our dataset is the inclusion of a ``user-preference'' white-balance ground truth that focuses on matching real-world scene observations and enhancing image aesthetics. Figure~\ref{fig:user_preference}-A shows the chromaticity distribution of both the neutral ground truth (obtained from the color chart) and the user-preference ground truth in our dataset. As shown, the neutral ground truth spans a larger area in the $rg$ chromaticity space, which is intuitive, as it represents the true color of the illuminant lighting the scene. In contrast, the user-preference ground truth has a narrower distribution near the Planckian locus. This explains the lower angular errors observed in most methods when compared to the neutral illuminant estimation results.

\subsection{Statistics}
\label{supp-sec:statistics}
Figure~\ref{fig:categories} shows the statistics of lighting classes (i.e., artificial lights such as incandescent and fluorescent, and natural lights such as outdoor daylight) and scene classes (daylight, sunset/sunrise, night, and indoor) in our dataset. The training, validation, and testing splits are evenly distributed across the different lighting and scene classes.

\begin{figure*}[!t]
\centering
\includegraphics[width=\linewidth]{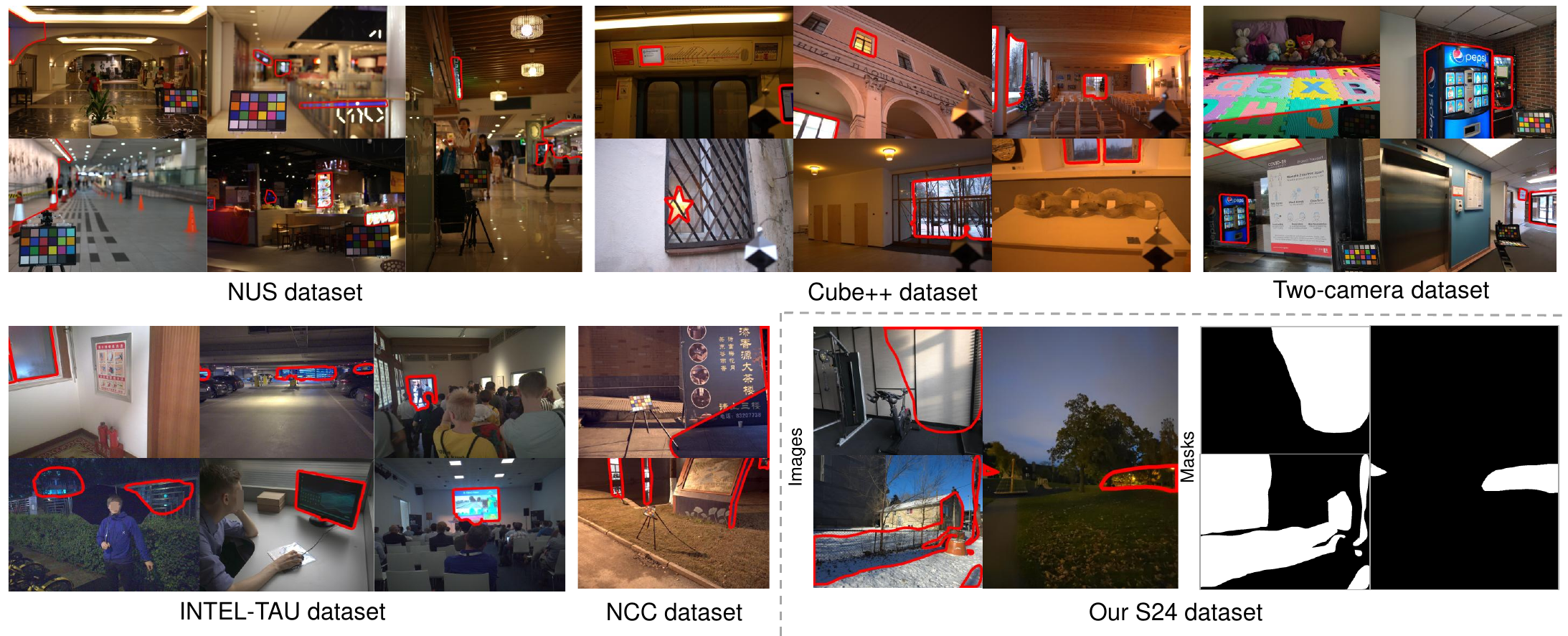}
\vspace{-4mm}
\caption{Unlike other color constancy datasets (e.g., NUS \cite{NUS}, Cube++ \cite{Cube++}, Two-camera \cite{two-camera}, INTEL-TAU \cite{IntelTAU}, NCC \cite{NCC}), our dataset provides masks for scenes lit by illuminants different from the dominant illuminant used as the ground-truth illuminant color. In this figure, we show examples from each dataset, including ours, where each scene contains regions (highlighted with red borders) that are either lit by or have source lighting different from the dominant illuminant color of the scene. For our dataset, we also show the corresponding manually created masks for the shown images. All shown images are in the sRGB color space. \label{fig:multi_illum_examples}}
\end{figure*}

\begin{figure}[t]
\centering
\includegraphics[width=\linewidth]{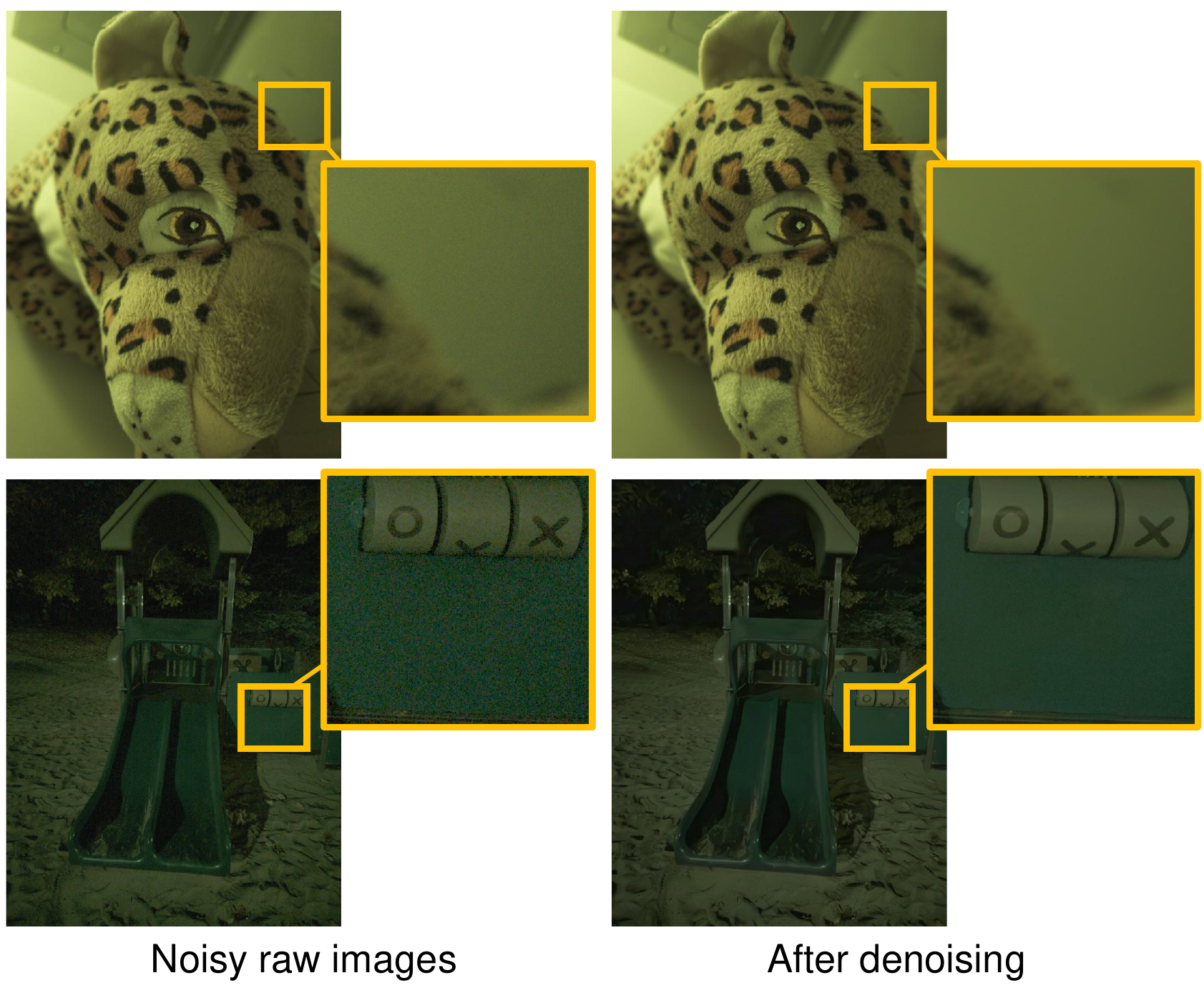}
\vspace{-4mm}
\caption{Our dataset includes denoised images that can serve as proxy ground truth for learnable denoisers. To illustrate the effect of denoising, we present raw images before and after denoising, with gamma correction applied for better visualization.\label{fig:denoising}}
\vspace{-2mm}
\end{figure}

\begin{figure*}[h]
\centering
\includegraphics[width=0.98\linewidth]{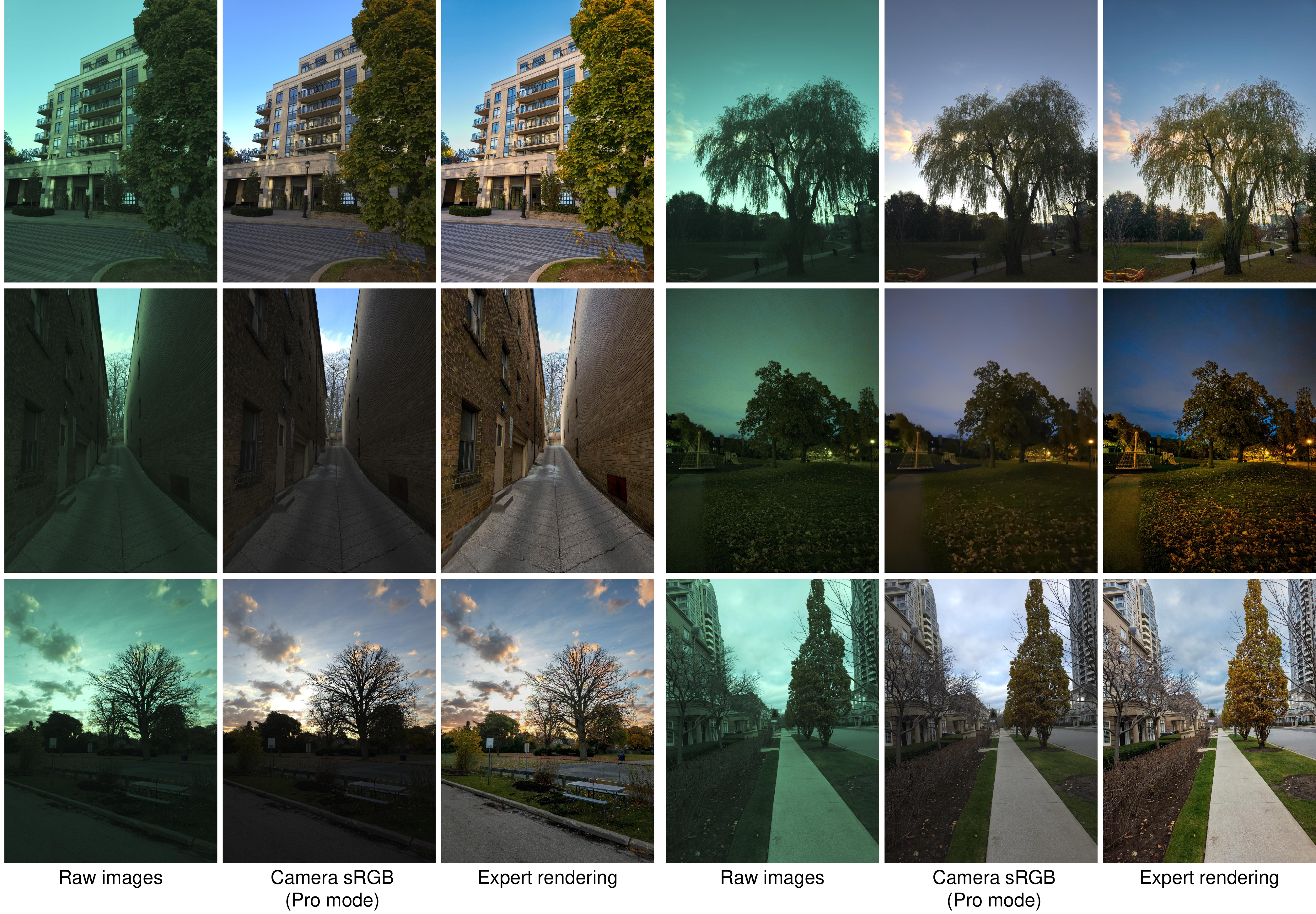}
\vspace{-1mm}
\caption{Our dataset includes sRGB images produced by the in-camera lightweight ISP and expert-rendered sRGB images from Adobe Lightroom, which incorporate local tone mapping adjustments to enhance aesthetic appeal.\label{fig:srgb}}
\end{figure*}

\subsection{Data labeling}
\label{supp-sec:gui-tool}

\begin{figure*}[t]
\centering
\includegraphics[width=\linewidth]{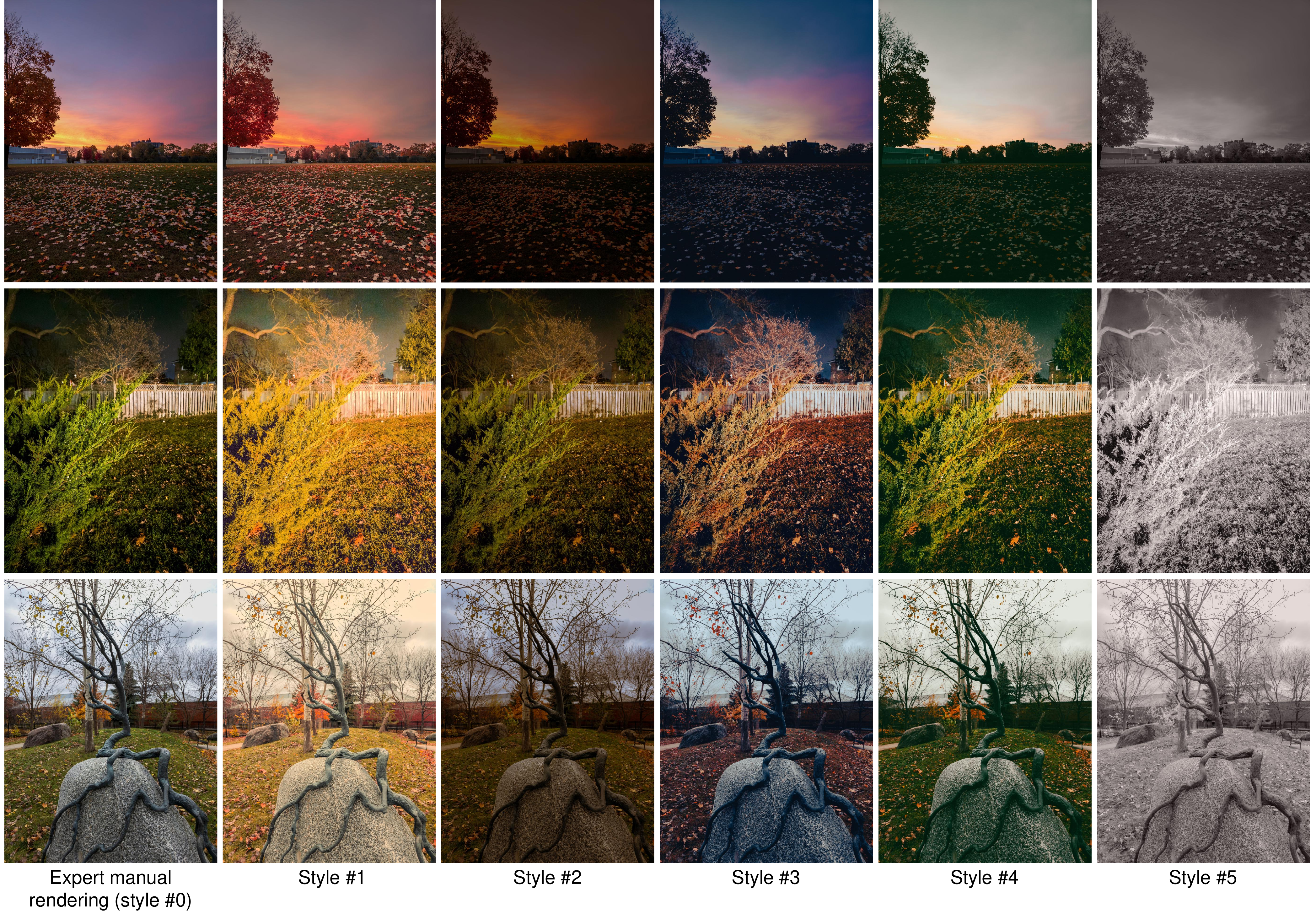}
\vspace{-4mm}
\caption{In addition to the expert rendering (style \#0), our dataset includes sRGB images with multiple styles, which can serve as ground truth for picture style transfer or raw-to-multi-style sRGB rendering.\label{fig:styles}}
\end{figure*}

To facilitate the annotation process, we developed a Matlab graphical user interface (GUI) tool; see Fig.~\ref{fig:gui}-A. An expert photographer was instructed to select a reference white point of the scene from the raw image of the color chart for each scene, copy it, and paste it to assign as the ground-truth neutral illuminant color for the sequential scene(s) sharing the same lighting condition.  In addition, the annotator was asked to assign a ``user-preference'' ground truth, which may not align with the neutral white-balance appearance or the in-camera white-balance result of the Samsung S24 Ultra; see Fig.~\ref{fig:user_preference}-B. Notably, the user-preference ground truth is intended to reflect real-world observations and enhance the scene's aesthetics, and therefore, may differ from both the neutral and camera-based ground truths. The mean angular error between the annotated user-preference ground truth and the neutral white-balance ground truth is 2.67$^\circ$, and the error between the user-preference and the illuminant colors from the in-camera AWB module is 1.34$^\circ$.

The user-preference tools allow the annotator to interpolate between the camera white balance setting (produced by the in-camera illuminant estimation method) and the annotated neutral white balance. Additionally, the annotator can adjust the user-preference white point to make the scene appear cooler or warmer by modifying the correlated color temperature (CCT).

To map between illuminant RGB colors in the camera raw space and CCTs, we captured a color chart under various CCTs ranging from 1,325K to 10,000K using a controllable light booth, see Fig.~\ref{fig:gui}-B. We then measured the raw RGB color corresponding to each CCT by manually selecting gray patches from the color chart and averaging them for each raw image. We then fit a linear regression model to map the $R/G$ and $B/G$ chromaticity values of raw white points to the corresponding CCT. To convert the CCT value back to the normalized RGB illuminant color, we locate the nearest CCT value within the calibrated CCTs. Subsequently, we linearly interpolate between the corresponding measured chroma values of the nearest lower and higher CCTs.

While this is a simplified method for converting between chromaticity values and CCTs, it was sufficient for our goal to enable the annotator to adjust the white balance in an interpretable manner. These adjustments could be achieved either by modifying the CCT or interpolating between the camera's white balance and the neutral white balance settings, rather than directly adjusting the RGB values of the illuminant, which can be more challenging to fine-tune for the desired results.

Our dataset includes a diverse range of scenes captured under various lighting conditions, including night scenes, making it challenging to ensure the presence of a single light source in each scene. To address this, we complemented white-balance labeling with binary masks for scenes containing multiple light sources. These masks identify regions illuminated by light sources different from the dominant light used to label the ground truth. See Fig.~\ref{fig:multi_illum_examples} for example masks.

Additionally, to ensure privacy, we applied blurring to personal information (e.g., faces, license plates, phone numbers, etc.) across both the raw and camera sRGB images in our dataset.

Each scene is further annotated with its scene class (daylight, sunset/sunrise, night, and indoor) and lighting condition class (artificial or natural light). Although these labels were primarily used for dataset statistics, we believe they hold significant potential for future research. For example, the scene class could be an additional input feature to improve model accuracy.

The GUI tool also facilitates the assignment of images to one of three sets: training, testing, or validation. The primary criterion was to ensure that testing and validation sets were distinct, containing no overlapping scenes with the training set. We further evaluated the testing and validation sets by applying the gray-world algorithm \cite{GW} to selected images, generating real-time statistics that provided insights into their complexity. Since the gray-world algorithm is a simple baseline, its angular error served as a useful indicator of the difficulty of these sets. Finally, we visually reviewed the testing and validation sets to ensure they comprised unique and diverse scenes.

\subsection{User study}
\label{supp-sec:user-study}
To validate the annotation of user-preference ground truth, we conducted a user study with 20 participants who had normal vision. We first sorted the images by the angular error between the user-preference ground truth and the neutral ground truth. 

From the images with the highest angular errors between the user-preference and neutral ground truths, we randomly selected 100 images. 

\begin{figure*}[!t]
\centering
\includegraphics[width=\linewidth]{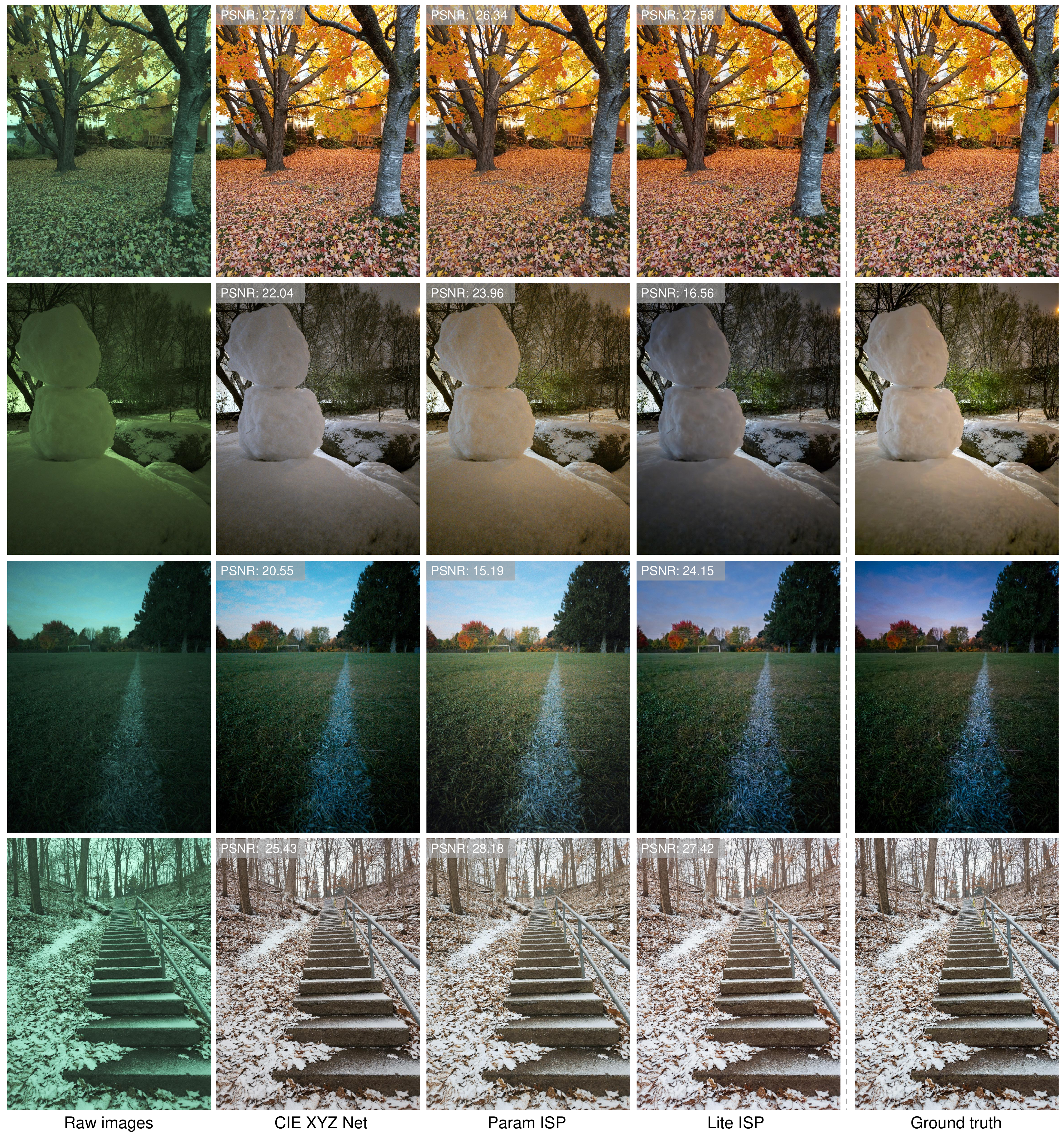}
\vspace{-4mm}
\caption{Qualitative results from our testing set for trained models rendering raw images to our expert-rendered sRGB. Results are shown for CIE XYZ Net \cite{afifi2021cie}, Param ISP \cite{kim2023paramisp}, and Lite ISP \cite{RAW-to-sRGB}. Raw images have undergone gamma correction for better visualization. \label{fig:isp}}
\end{figure*}

\begin{figure*}[!t]
\centering
\includegraphics[width=0.95\linewidth]{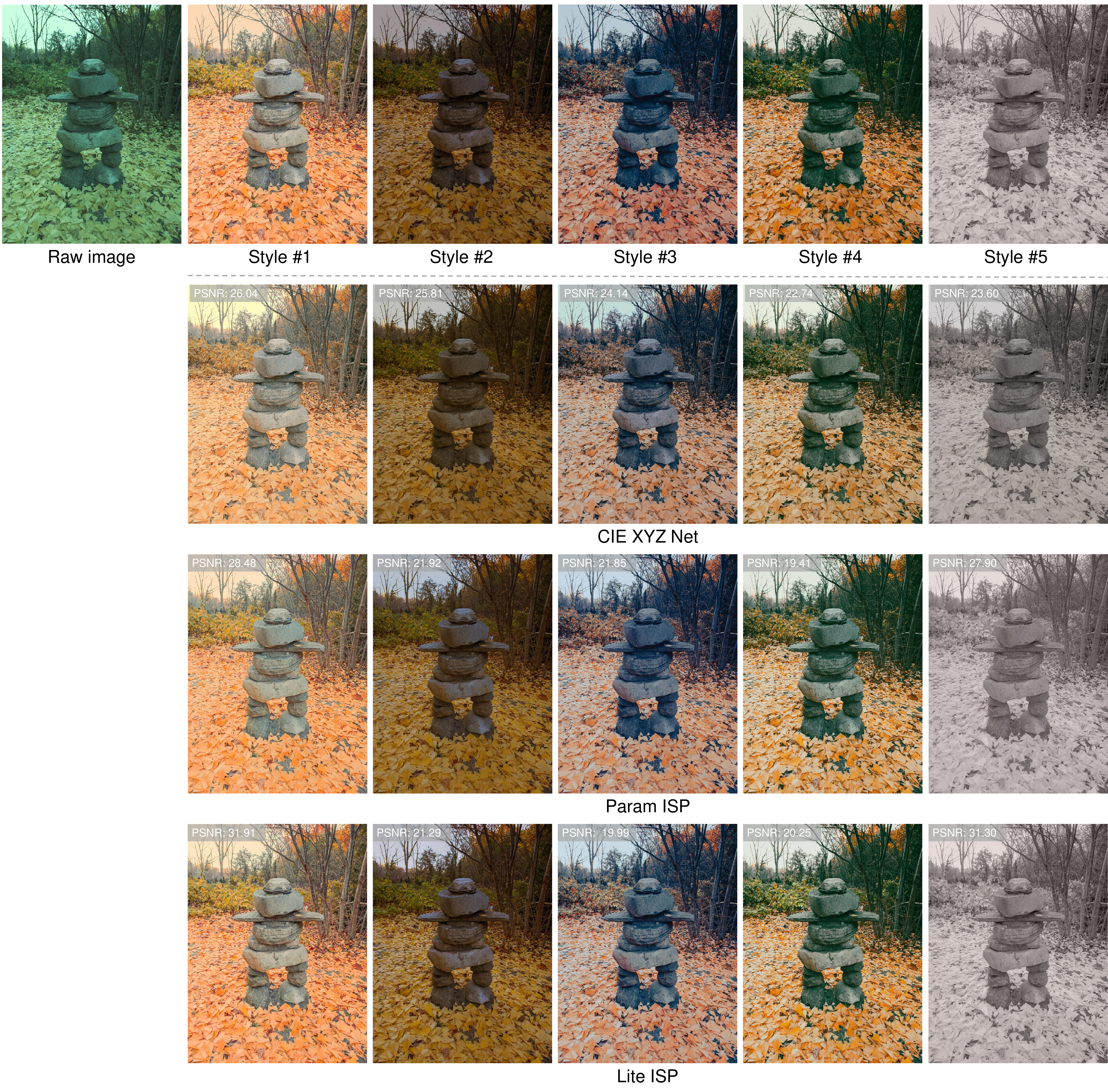}
\vspace{-2mm}
\caption{Qualitative results from our testing set for trained models rendering raw images to the sRGB ground truth of our five additional picture styles. The top row shows the raw image (gamma corrected for visualization) and the corresponding ground-truth image with different picture styles. The second row presents results from CIE XYZ Net \cite{afifi2021cie}, Param ISP \cite{kim2023paramisp}, and Lite ISP \cite{RAW-to-sRGB}. \label{fig:isp-style}}
\end{figure*}

For each participant, we performed 100 trials, showing these 100 images one by one. In each trial, we present the participant with two versions of white-balanced images corresponding to each ground truth, after applying the color correction matrices and gamma correction for better visualization on a calibrated monitor.

Participants were asked to select the image that appeared most natural. To provide context, we also showed the time of day at which the image was captured (e.g., sunset, sunrise, daylight, etc.). Overall, participants selected the user-preference ground truth in 71.95\% of the trials, and selected the neutral ground truth in 28.05\% of the trials, confirming that the user-preference ground truth is preferred by users most of the time. 

\section{Additional ground truth and applications}
\label{sec:additional-gt-data}

\paragraph{Denoising:}
We used Adobe Lightroom AI denoiser to generate denoised raw images that simulate in-camera denoising. These denoised images were used to compute noise stats, $\mathbf{n}$, which served as one of the input features for our model. The Adobe Lightroom AI denoiser may leverage camera-specific information to achieve effective denoising in the linear raw space, requiring minimal manual adjustment to produce satisfactory results.

These denoised raw images, included as additional ``ground-truth'' data in our dataset, can serve as a proxy for evaluating denoising algorithms (e.g., \cite{wang2020practical, liu2021invertible}) in the linear raw space across diverse lighting conditions, including dark scenes; see Fig.~\ref{fig:denoising}.

\paragraph{Expert sRGB rendering:}
Our dataset was captured in Samsung Pro mode using the Samsung S24 Ultra, providing pre-processed raw images after early-stage operations such as demosaicing. The device also produces an sRGB image by rendering the raw image through a simplified version of the camera's native ISP, which lacks accurate local tone mapping and denoising.

To improve sRGB rendering, we manually processed the raw images in Adobe Lightroom, including local tone mapping adjustments. Specifically, an expert photographer rendered all 3,224 denoised raw images to sRGB, enhancing their aesthetic appeal through global and local tone mapping adjustments using manually created spatial masks in Adobe Lightroom. Although our rendering approach may seem similar to that of the MIT-Adobe FiveK dataset \cite{fivek}, which also involves expert manual rendering using Adobe Lightroom, our dataset, captured with the more recent Samsung S24 Ultra, provides a more up-to-date representation than the older DSLR cameras used in Adobe FiveK. Moreover, Adobe FiveK lacks local tonal adjustments in expert rendering. Our rendered sRGB images leverage the latest denoising techniques in Adobe Lightroom and its advanced functionality to achieve high-quality tone mapping, including local tone adjustments (see Fig.~\ref{fig:srgb}).

These high-quality rendered sRGB images make our dataset a valuable resource for the raw-to-sRGB rendering task. In contrast to existing raw-to-sRGB datasets (e.g., the Zurich raw-to-RGB dataset \cite{ignatov2019aim} and the Samsung S7 dataset \cite{schwartz2018deepisp}), which suffer from input–ground truth misalignment \cite{ignatov2019aim} or contain limited numbers of images (e.g., fewer than 250 full-resolution images \cite{ignatov2019aim, schwartz2018deepisp}) with restricted scene diversity and lighting conditions (e.g., primarily daylight \cite{ignatov2019aim}), our dataset offers well-aligned, high-resolution (4000$\times$3000) raw, denoised raw, and sRGB ground-truth images across diverse scenes and lighting conditions. This makes it a reliable resource for training neural ISP methods aimed at rendering raw images into high-quality sRGB images (e.g., \cite{ignatov2022microisp, kim2023paramisp, he2024enhancing}).


\begin{table}[!t]
\centering
\caption{Quantitative results for rendering raw images to expert-rendered sRGB on our test set. Each method was trained on our training set to map raw images to expert-rendered sRGB. \label{tab:neural-isp}}\vspace{-1mm}
\resizebox{0.48\textwidth}{!}{
\begin{tabular}{l|ccccc}

\textbf{Method} & \textbf{PSNR}  & \textbf{SSIM}  & \textbf{LPIPS} &  $\mathbf{\Delta E_{2000}}$ &  \begin{tabular}[c]{@{}c@{}}\textbf{\#params} \\ \textbf{(K)}\end{tabular} \\ \hline
 
CIE XYZ Net \cite{afifi2021cie} & 23.32 & 0.8596 & 0.1242 & 7.0239 & 1,348.8\\

Invertible ISP \cite{xing2021invertible} & 22.87 & 0.8197 & 0.1468 & 7.3739 & 1,413.8 \\

Param ISP \cite{kim2023paramisp} & 24.32 & 0.8411 & 0.1145 & 6.1353 & 1,420.0\\

Lite ISP \cite{RAW-to-sRGB} & 25.49 & 0.8967 & 0.0744 & 5.5213 & 9,094.0\\

Fourier ISP \cite{he2024enhancing} & 24.50 & 0.9125 & 0.0962 & 5.9276 & 7,589.8\\

\end{tabular}
}
\end{table}


\begin{table*}[!h]
    \centering
    \caption{Quantitative results for rendering raw images to five different styles in sRGB. Each method was trained on our training set to map raw images to sRGB images rendered in a specific style. \label{tab:neural-isp-styles}}\vspace{-1mm}
    \scalebox{0.62}{
     \addtolength{\tabcolsep}{-0.2em}
     \begin{tabular}{c|cccc|cccc|cccc|cccc|cccc}
    \multirow{2}{*}{\textbf{Method}} & \multicolumn{4}{c|}{\textbf{Style 1}} & \multicolumn{4}{c|}{\textbf{Style 2}} & \multicolumn{4}{c|}{\textbf{Style 3}} & \multicolumn{4}{c|}{\textbf{Style 4}} & \multicolumn{4}{c}{\textbf{Style 5}} \\ \cline{2-21} 
                            & \multicolumn{1}{c}{PSNR} & \multicolumn{1}{c}{SSIM} & \multicolumn{1}{c}{LPIPS} & \multicolumn{1}{c|}{$\Delta E_{2000}$} & \multicolumn{1}{c}{PSNR} & \multicolumn{1}{c}{SSIM} & \multicolumn{1}{c}{LPIPS} & \multicolumn{1}{c|}{$\Delta E_{2000}$} & \multicolumn{1}{c}{PSNR} & \multicolumn{1}{c}{SSIM} & \multicolumn{1}{c}{LPIPS} & \multicolumn{1}{c|}{$\Delta E_{2000}$} & \multicolumn{1}{c}{PSNR} & \multicolumn{1}{c}{SSIM} & \multicolumn{1}{c}{LPIPS} & \multicolumn{1}{c|}{$\Delta E_{2000}$} & \multicolumn{1}{c}{PSNR} & \multicolumn{1}{c}{SSIM} & \multicolumn{1}{c}{LPIPS} & \multicolumn{1}{c}{$\Delta E_{2000}$} \\ \hline
    CIE XYZ Net \cite{afifi2021cie} & 22.40 & 0.8586 & 0.1762 & 8.2912 & 24.05 & 0.8720 & 0.1199 & 6.6336 & 22.00 & 0.8536 & 0.1615 & 8.9058 & 22.26 & 0.8457 & 0.1468 & 7.6241 & 24.67 & 0.8967 & 0.1326 & 5.2699 \\
    Invertible ISP \cite{xing2021invertible} & 23.48 & 0.8322 & 0.1571 & 6.6418 & 26.35 & 0.8606 & 0.115 & 5.2289 & 23.84 & 0.8518 & 0.1437 & 7.3474 & 23.33 & 0.8422 & 0.1445 & 7.6935 & 24.9 & 0.8748 & 0.1626 & 5.9271  \\
    Param ISP \cite{kim2023paramisp} & 24.97 & 0.8559 & 0.123 & 5.7238 & 27.81 & 0.8750 & 0.0952 & 4.9218 & 27.11 & 0.869 & 0.1005 & 4.9472 & 24.18 & 0.8533 & 0.1184 & 6.1376 & 25.43 & 0.8667 & 0.1344 & 4.4993 \\
    Lite ISP \cite{RAW-to-sRGB} & 26.66 & 0.9145 & 0.0668 & 4.7019 & 28.33 & 0.9224 & 0.0636 & 4.2839 & 26.31 & 0.9126 & 0.0729 & 5.2200 & 25.04 & 0.8942 & 0.082 & 5.5165 & 28.07 & 0.9353 & 0.0707 & 3.4339 \\
    Fourier ISP \cite{he2024enhancing} & 25.19 & 0.925 & 0.0985 & 5.432 & 28.03 & 0.9276 & 0.0819 & 4.4879 & 25.38 & 0.9186 & 0.0997 & 5.7031 & 24.74 & 0.9063 & 0.0996 & 5.5908 & 27.41 & 0.9468 & 0.0889 & 3.5606 \\
\end{tabular}
}
\end{table*}


\paragraph{sRGB picture styles:}

In addition to the sRGB images from the Samsung S24 Ultra (Pro mode) and our expert-rendered sRGB images, we provide five additional sRGB versions for each raw image using Adobe Lightroom presets, similar to \cite{elezabi2024inretouch}. These can serve as ground truth for picture style transfer or raw-to-multiple-style sRGB rendering. See Fig.~\ref{fig:styles}.

\paragraph{Results of raw-to-sRGB rendering:} We evaluate different neural ISP methods that aim to render raw images into corresponding sRGB images using our dataset, which includes expert-rendered sRGB ground truth and five additional picture styles. Specifically, we trained the methods in \cite{afifi2021cie, xing2021invertible, kim2023paramisp, RAW-to-sRGB, he2024enhancing} on our training set to map noisy raw images to expert-rendered sRGB. Additionally, we trained each method to map raw images to each of our five picture styles.  Table~\ref{tab:neural-isp} shows PSNR, SSIM \cite{wang2004image}, LPIPS \cite{zhang2018unreasonable}, and $\Delta E_{2000}$ \cite{sharma2005ciede2000} results on our test set for the evaluated neural ISP methods. We also report results for the five different styles, where each model was trained to map raw images to a specific target style in the sRGB space, as shown in Table~\ref{tab:neural-isp-styles}. Figures~\ref{fig:isp} and \ref{fig:isp-style} provide qualitative examples comparing these methods’ outputs with the ground-truth images.

{
    \small
    \bibliographystyle{ieeenat_fullname}
    \bibliography{ref}
}


\end{document}

%% file: tables/results_test_wo_mask_w_params.tex
\begin{table*}[t]
\centering
\caption{\textbf{Results on the testing set}. We report the mean, median, best 25\%, worst 25\%, tri-mean, and maximum angular errors for each method on neutral and user-preference white-balance ground-truth illuminants, presented in the format (neutral / user-preference). Symbols $\mathbf{n}$ and $\mathbf{r}$ represent noise stats and SNR stats, respectively. The \colorbox{best}{best} and \colorbox{secondbest}{second-best} results are highlighted. \label{tab:results-test-wo-mask}}\vspace{-1mm}
\scalebox{0.79}{
\begin{tabular}{l|lllllll|l}
\textbf{Method} & \textbf{Mean}  & \textbf{Med.}  & \begin{tabular}[c]{@{}c@{}}\textbf{Best} \\ \textbf{25\%}\end{tabular} & \begin{tabular}[c]{@{}c@{}}\textbf{Worst} \\ \textbf{25\%}\end{tabular} & \begin{tabular}[c]{@{}c@{}}\textbf{Worst} \\ \textbf{5\%}\end{tabular} & \textbf{Tri.}   & \textbf{Max} &  \begin{tabular}[c]{@{}c@{}}\textbf{\#params} \\ \textbf{(K)}\end{tabular} \\ \hline

GW \cite{GW}   & 6.23 / 5.54 & 6.01 / 4.52 & 1.02 / 1.00 & 12.43 / 11.91 & 18.05 / 19.81 & 5.76 / 4.65 & 28.09 / 31.70 &   - \\ 
SoG \cite{SoG}   & 4.45 / 3.59 & 3.54 / 2.17 & 0.74 / 0.67 & 9.61 / 8.81 & 14.94 / 15.61 & 3.77 / 2.62 & 23.22 / 29.36 &  - \\ 
GE-1st \cite{GE}   & 4.21 / 3.33 & 3.29 / 2.12 & 0.71 / 0.55 & 9.34 / 8.29 & 15.34 / 15.19 & 3.49 / 2.36 & 27.81 / 28.80 &  - \\ 
GE-2nd \cite{GE}   & 4.11 / 3.18 & 3.17 / 1.89 & 0.70 / 0.58 & 9.09 / 7.79 & 14.90 / 14.85 & 3.35 / 2.18 & 25.09 / 29.42 &  - \\ 
Max-RGB \cite{maxRGB}  & 4.01 / 2.61 & 2.92 / 1.88 & 1.06 / 0.97 & 8.57 / 5.69 & 14.01 / 11.07 & 3.30 / 1.94 & 34.22 / 23.63 &  - \\
wGE \cite{wGE}   & 3.96 / 3.07 & 2.96 / 1.91 & 0.62 / 0.49 & 8.95 / 7.79 & 14.94 / 15.08 & 3.21 / 2.10 & 31.33 / 30.54 &  - \\ 
PCA \cite{NUS}   & 4.42 / 3.63 & 3.54 / 1.95 & 0.70 / 0.61 & 9.59 / 9.19 & 14.46 / 16.16 & 3.76 / 2.52 & 22.87 / 31.58 &  - \\  
MSGP \cite{MSGP}   & 6.39 / 5.69 & 5.72 / 4.38 & 0.94 / 0.98 & 13.39 / 12.73 & 20.97 / 22.33 & 5.64 / 4.68 & 37.25 / 36.92 &  - \\  
GI \cite{GI}   & 4.70 / 4.93 & 3.19 / 2.97 & 0.44 / 0.78 & 11.63 / 12.22 & 20.48 / 21.35 & 3.48 / 3.53 & 36.34 / 36.02 &  - \\  
TECC \cite{TECC}   & 4.12 / 3.17 & 3.23 / 1.91 & 0.74 / 0.55 & 9.10 / 7.83 & 14.73 / 14.48 & 3.46 / 2.17 & 27.08 / 28.87 &  - \\  
Gamut (pixels) \cite{GAMUT}   & 3.77 / 2.40 & 2.81 / 1.49 & 0.77 / 0.63 & 8.31 / 5.84 & 12.93 / 11.18 & 3.16 / 1.63 & 21.53 / 23.31 &  0.636 \\ 
Gamut (edges) \cite{GAMUT}   & 4.45 / 3.94 & 3.52 / 3.04 & 1.08 / 1.01 & 9.51 / 8.50 & 15.00 / 15.25 & 3.70 / 3.18 & 28.55 / 29.46 &  324 \\ 
Gamut (1st) \cite{GAMUT}   & 4.10 / 3.65 & 3.08 / 2.67 & 0.72 / 0.90 & 9.26 / 8.25 & 14.75 / 14.69 & 3.28 / 2.91 & 21.38 / 26.35 &  279 \\ 
NIS \cite{NIS}   & 4.58 / 3.90 & 3.76 / 2.57 & 0.77 / 0.75 & 9.81 / 9.10 & 14.70 / 15.66 & 3.87 / 2.97 & 20.76 / 31.80 &  0.078 \\  
Classification-CC \cite{CLASSIFICATION}   & 2.73 / 1.61 & 1.98 / 1.16 & 0.61 / 0.36 & 6.03 / 3.60 & 9.37 / 6.07 & 2.18 / 1.27 & 19.34 / 9.53 &   58,384 \\ 
FFCC \cite{FFCC}   & 2.62 / 1.50 & 1.46 / 0.81 & 0.37 / 0.24 & 6.89 / 3.99 & 16.59 / 8.43 & 1.66 / 0.95 & 48.97 / 18.60 &  12 \\ 
FFCC (capture info) \cite{FFCC}   & 2.31 / 1.35 & 1.38 / 0.80 & \colorbox{secondbest}{0.34} / 0.23 & 5.82 / 3.51 & 12.44 / 7.36 & 1.60 / 0.88 & 47.67 / 16.96 &  36.9 \\
FC4 \cite{FC4}   & 3.80 / 2.65 & 2.78 / 2.25 & 0.85 / 0.85 & 8.61 / 5.14 & 15.06 / 7.52 & 2.86 / 2.37 & 25.74 / 11.42 &  1,705 \\ 
APAP (GW) \cite{APAP}   & 3.74 / 2.09 & 3.14 / 1.67 & 0.93 / 0.49 & 7.72 / 4.44 & 11.09 / 6.89 & 3.26 / 1.76 & \colorbox{secondbest}{16.43} / \colorbox{best}{9.02} &  0.289 \\ 
SIIE \cite{SIIE}  & 4.09 / - & 3.25 / - & 0.91 / - & 8.97 / - & 15.96 / - & 3.37 / - & 43.24 / - &  1,008 \\ 
SIIE (tuned) \cite{SIIE}  & 3.15 / - & 2.22 / - & 0.51 / - & 7.28 / - & 12.14 / - & 2.46 / - & 34.52 / - &  1,008 \\ 
SIIE (tuned-CS) \cite{SIIE}  & 3.14 / 1.74 & 2.20 / 1.20 & 0.50 / 0.32 & 7.39 / 4.06 & 13.61 / 6.73 & 2.41 / 1.30 & 38.96 / 9.44 &  1,008 \\ 
KNN (raw) \cite{knn}   & 2.44 / 1.41 & 1.51 / 0.83 & 0.36 / \colorbox{secondbest}{0.20} & 6.13 / 3.66 & 11.64 / 7.01 & 1.66 / 0.93 & 28.95 / 13.49 &  757 \\ 
Quasi-U-CC \cite{QUASI-CC}   & 3.85 / 3.26 & 2.97 / 1.81 & 0.55 / 0.58 & 8.47 / 8.24 & 13.30 / 15.78 & 3.25 / 2.21 & 24.74 / 32.17 &  54,421 \\ 
Quasi-U-CC (tuned) \cite{QUASI-CC}   & 3.11 / 2.54 & 2.27 / 1.43 & 0.49 / 0.44 & 7.14 / 6.57 & 11.53 / 13.62 & 2.44 / 1.63 & 22.67 / 33.74 &  54,421 \\ 
BoCF \cite{BoCF}   & 3.54 / 2.14 & 2.68 / 1.59 & 0.96 / 0.48 & 7.31 / 4.74 & 11.40 / 7.85 & 2.96 / 1.72 & 22.31 / 19.68 &  59  \\ 
C4 \cite{C4}   & 1.92 / 1.49 & 1.30 / 0.90 & 0.36 / 0.24 & 4.64 / 3.82 & 9.08 / 7.53 & 1.40 / 1.03 & 21.55 / 18.09 &  5,116 \\ 
CWCC \cite{CWCC}   & 3.65 / 2.30 & 2.71 / 1.72 & 0.82 / 0.67 & 7.96 / 4.95 & 12.52 / 9.65 & 2.99 / 1.81 & 18.66 / 20.81 &  101 \\ 
C5 \cite{C5}   & 3.22 / - & 2.51 / - & 0.78 / - & 6.97 / - & 10.54 / - & 2.68 / - & \colorbox{best}{16.38} / - &  412 \\ 
C5 (tuned) \cite{C5}   & 1.91 / - & \colorbox{secondbest}{1.24} / - & 0.38 / - & 4.57 / - & \colorbox{secondbest}{8.43} / - & 1.38 / - & 17.22 / - &  412 \\ 
C5 (tuned-CS) \cite{C5}   & 1.95 / 1.25 & 1.32 / 0.84 & 0.37 / 0.23 & 4.72 / \colorbox{best}{2.94} & \colorbox{best}{8.15} / \colorbox{best}{4.98} & 1.44 / 0.93 & 16.78 / \colorbox{secondbest}{9.23} &  172 \\ 
TLCC \cite{TLCC}   & 2.71 / 2.74 & 2.06 / 1.83 & 0.66 / 0.69 & 5.89 / 6.30 & 10.30 / 12.90 & 2.17 / 1.99 & 21.44 / 33.33 &  32,910 \\ 
PCC \cite{PCC}    & 3.03 / 1.67 & 2.13 / 1.20 & 0.53 / 0.40 & 7.08 / 3.79 & 11.20 / 6.84 & 2.34 / 1.27 & 16.82 / 12.33 &  0.378 \\ 
RGP \cite{RGP}  & 4.59 / 4.56 & 3.13 / 2.81 & 0.43 / 0.70 & 11.13 / 11.27 & 18.79 / 20.25 & 3.53 / 3.31 & 32.11 / 33.98 &  - \\
CFCC \cite{CFCC}  & 3.07 / 1.54 & 2.20 / 1.05 & 0.73 / 0.39 & 6.87 / 3.55 & 12.55 / 6.98 & 2.36 / 1.13 & 23.34 / 14.40 &  0.283 \\ \hdashline
Ours (w/o $\mathbf{n}$, w/o $\mathbf{r}$) & 1.93 / 1.26 & 1.35 / 0.77 & 0.38 / 0.23 & \colorbox{secondbest}{4.56} / 3.13 & 9.48 / 5.88 & 1.43 / 0.88 & 22.63 / 15.90 & 4.83 \\ 
Ours (w/o $\mathbf{n}$, w/ $\mathbf{r}$)   & 1.89 / \colorbox{secondbest}{1.23} & \colorbox{best}{1.18} / 0.79 & \colorbox{best}{0.32} / 0.24 & 4.74 / 3.01 & 10.10 / \colorbox{secondbest}{5.07} & \colorbox{best}{1.30} / 0.90 & 24.99 / 12.91 &  4.93 \\
Ours (w/ $\mathbf{n}$, w/o $\mathbf{r}$)   & \colorbox{secondbest}{1.87} / \colorbox{best}{1.20} & \colorbox{secondbest}{1.24} / \colorbox{best}{0.72} & 0.37 / 0.24 & 4.58 / 2.99 & 9.82 / 5.58 & \colorbox{best}{1.30} / \colorbox{best}{0.82} & 29.46 / 15.12 &  4.93 \\
Ours (w/ $\mathbf{n}$, w/ $\mathbf{r}$)   & \colorbox{best}{1.84} / \colorbox{best}{1.20} & \colorbox{secondbest}{1.24} / \colorbox{secondbest}{0.77} & 0.35 / \colorbox{best}{0.19} & \colorbox{best}{4.41} / \colorbox{secondbest}{2.95} & 9.17 / 5.12 & \colorbox{secondbest}{1.32} / \colorbox{secondbest}{0.87} & 35.42 / 12.71 &  5.03 \\

\end{tabular}
}
\end{table*}

%% file: tables/results_cube.tex
\begin{table}[t]
\centering
\caption{\textbf{Results on the testing set of the `Simple Cube++' dataset \cite{Cube++}}. We report the mean, median, best 25\%, worst 25\%, tri-mean, and maximum angular errors for each method. Symbols $\mathbf{c}$, $i$, $e$, and $\mathbf{r}$ refer to the time-capture feature, ISO, exposure time, and SNR stats, respectively. The \colorbox{best}{best} and \colorbox{secondbest}{second-best} results are highlighted. \label{tab:results-cube}}\vspace{-1mm}
\scalebox{0.65}{
\begin{tabular}{l|ccccccc}
\textbf{Method} & \textbf{Mean}  & \textbf{Med.}  & \begin{tabular}[c]{@{}c@{}}\textbf{Best} \\ \textbf{25\%}\end{tabular} & \begin{tabular}[c]{@{}c@{}}\textbf{Worst} \\ \textbf{25\%}\end{tabular} & \begin{tabular}[c]{@{}c@{}}\textbf{Worst} \\ \textbf{5\%}\end{tabular} & \textbf{Tri.}   & \textbf{Max} \\ \hline

FFCC \cite{FFCC}  & 1.38 & \colorbox{best}{0.58} & \colorbox{secondbest}{0.18} & 4.05 & 10.30 & \colorbox{secondbest}{0.68} & 45.20  \\   
FC4 \cite{FC4}   & 3.72 & 2.02 & 0.47 & 9.93 & 17.18 & 2.40 & 29.92 \\ 
C4 \cite{C4}    & 1.16 & 0.65 & 0.24 & 3.00 & 6.72 & 0.73 & 13.95\\  
C5 (tuned-CS) \cite{C5}   & 1.19 & 0.60 & 0.18 & 3.26 & 7.61 & \colorbox{secondbest}{0.68} & 15.62 \\  
TLCC \cite{TLCC}  & 1.82 & 1.15 & 0.39 & 4.49 & 9.63 & 1.20 & 19.02 \\   \hdashline 

Ours ($\mathbf{c}=[i, e]^T$) & \colorbox{secondbest}{1.08}  & \colorbox{secondbest}{0.59} & \colorbox{best}{0.17} & \colorbox{secondbest}{2.84} & \colorbox{best}{5.53} & 0.69  & \colorbox{secondbest}{11.03}  \\ 
Ours ($\mathbf{c}=[i, e, \mathbf{r}^T]^T$) & \colorbox{best}{1.01} & 0.60 & \colorbox{best}{0.17} & \colorbox{best}{2.68} & \colorbox{secondbest}{5.71}  & \colorbox{best}{0.65} & \colorbox{best}{10.89} \\

\end{tabular}
}
\end{table}

%% file: tables/processing_time.tex
\begin{table}[!t]
\vspace{-5mm}
\centering
\caption{Processing time on an AMD Ryzen Threadripper PRO 3975WX CPU460
and an NVIDIA RTX A6000 GPU.  \label{tab:processing-time}}\vspace{-1mm}
\scalebox{0.76}{
\begin{tabular}{l|ccc}

\textbf{Method} & \textbf{CPU (ms)}  & \textbf{GPU (ms)}  & \textbf{FLOPs} \\ \hline


C4 \cite{C4} & 49.03 & 11.28 & 2.28G \\  
C5 \cite{C5} & 8.28 & 4.50 & 103.54M \\
\hdashline 
Ours & \colorbox{best}{0.55} & \colorbox{best}{0.45} & \colorbox{best}{16.78M} \\ 

\end{tabular}
}\vspace{-2mm}
\end{table}

%% file: tables/val_ablation.tex
\begin{table}[t]
    \centering
    \caption{\textbf{Results of ablation studies on the validation set} with neutral white-balance ground truth. We report the mean, median, best 25\%, and worst 25\% angular errors for our method with various configurations. Symbols $\mathbf{H}$, $\mathbf{p}$, $\mathbf{n}$, and $\mathbf{r}$ denote the histogram feature, time feature, noise stats, and SNR stats, respectively. The symbol $\mathbf{m}$ refers to the capture information feature, which includes ISO ($i$), shutter speed ($s$), and flash status ($f$), all of which are used in our input time-capture feature $\mathbf{c}$. The \colorbox{best}{best} results are highlighted. \label{tab:results-validation-ablation}}\vspace{-3mm}
    \scalebox{0.75}{
    \begin{tabular}{ccccc|ccccc} $\mathbf{H}$ & $\mathbf{m}$ & $\mathbf{p}$ & $\mathbf{n}$ & $\mathbf{r}$ 
    & \textbf{Mean}  & \textbf{Med.}  & \begin{tabular}[c]{@{}c@{}}\textbf{Best} \\ \textbf{25\%}\end{tabular} & \begin{tabular}[c]{@{}c@{}}\textbf{Worst} \\ \textbf{25\%}\end{tabular} & \begin{tabular}[c]{@{}c@{}}\textbf{Worst} \\ \textbf{5\%}\end{tabular}  \\ \hline
    
    \cmark & \xmark & \xmark & \xmark & \xmark & 2.03 & 1.36 & 0.31 & 4.90 & 8.90  \\
    
    \xmark & \cmark & \cmark & \cmark & \cmark   & 2.29 & 1.81 & 0.50 & 4.95 & 8.58  \\ 
    
    \xmark & \xmark & \cmark & \cmark & \xmark  & 2.37 & 1.59 & 0.45 & 5.48 & 9.03  \\
    
    \cmark & \xmark & \cmark & \cmark & \xmark  &  1.75 & 1.17 & \colorbox{best}{0.29} & 4.13 &  7.26  \\ 
    
    \cmark & \cmark & \xmark & \cmark & \cmark  & 1.89 & 1.21 & 0.31 & 4.60 & 9.41 \\ 
    
    \cmark & \cmark & \cmark & \xmark & \xmark  & 1.85 & 1.32  & 0.35  & 4.26 & 7.53  \\ 
    
    \cmark & \cmark & \cmark & \xmark & \cmark  & 1.72 & 1.16  & 0.30 & 4.13  & 8.16  \\
    
    \cmark & \cmark & \cmark & \cmark & \xmark   & 1.67 & \colorbox{best}{1.07} & \colorbox{best}{0.29} & 4.04  & 7.90   \\
    
    \cmark & \cmark & \cmark & \cmark & \cmark  &  \colorbox{best}{1.66}  & 1.20  & 0.33  & \colorbox{best}{3.77}  & \colorbox{best}{6.95}  \\
    
    \end{tabular}
    }
    \vspace{-3mm}
    \end{table}

%% file: tables/results_val_w_mask.tex
\begin{table*}[t]
\centering
\caption{\textbf{Results on the validation set with masking}. We report the mean, median, best 25\%, worst 25\%, tri-mean, and maximum angular errors for each method on both neutral and user-preference white-balance ground truth, presented in the format (neutral 
/ user-preference). Results for our method with various configurations are included, where $\mathbf{p}$, $i$, $s$, $f$, $\mathbf{n}$, and $\mathbf{r}$ represent the time feature, ISO, shutter speed, flash status, noise stats, and SNR stats, respectively. Symbols $\mathbf{H}$ and $\mathbf{c}$ denote histogram and time-capture feature. $\mathbf{H} \rightarrow \mathbf{I}_\texttt{chroma}$ indicates using R/G and B/G images instead of histograms, while $\mathbf{H}_{e}$ refers to histogram of image edges. $log$-$\mathbf{H}$ refers to the histogram used in \cite{CCC, C5} and $\mathbf{c}$-raw refers to using time-capture features without any pre-processing or normalization. Additional configurations include $h$ (number of histogram bins), and $uv$ coord. (additional histogram channels of the u/v coordinates in histogram space). The number of parameters required by each method is reported. The \colorbox{best}{best} and \colorbox{secondbest}{second-best} results are highlighted. \label{tab:results-validation}}\vspace{-1mm}
\scalebox{0.69}{
\begin{tabular}{l|ccccccc|c}
\textbf{Method} & \textbf{Mean}     & \textbf{Med.}                 & \begin{tabular}[c]{@{}c@{}}\textbf{Best} \\ \textbf{25\%}\end{tabular} & \begin{tabular}[c]{@{}c@{}}\textbf{Worst} \\ \textbf{25\%}\end{tabular} & \begin{tabular}[c]{@{}c@{}}\textbf{Worst} \\ \textbf{5\%}\end{tabular} & \textbf{Tri.}                 & \textbf{Max}    &  \begin{tabular}[c]{@{}c@{}}\textbf{\#params} \\ \textbf{(K)}\end{tabular} \\ \hline

GW \cite{GW}  & 5.86 / 5.41 & 4.90 / 3.97 & 0.78 / 0.95 & 12.74 / 11.92 & 19.86 / 19.06 & 4.98 / 4.42 & 30.05 / 30.53 &   -  \\
SoG \cite{SoG}  & 4.38 / 3.95 & 3.28 / 2.71 & 0.57 / 0.69 & 9.87 / 9.03 & 15.63 / 14.53 & 3.62 / 3.09 & 31.28 / 31.66 &  -  \\
GE-1st \cite{GE}  & 4.02 / 3.62 & 2.82 / 2.43 & 0.60 / 0.64 & 9.09 / 8.46 & 14.99 / 14.10 & 3.18 / 2.73 & 32.57 / 32.91 &  -  \\
GE-2nd \cite{GE}  & 3.71 / 3.29 & 2.67 / 2.29 & 0.62 / 0.60 & 8.29 / 7.56 & 14.15 / 13.17 & 2.96 / 2.52 & 31.66 / 32.02 &  -  \\
Max-RGB \cite{maxRGB}  & 3.54 / 2.57 & 2.75 / 1.84 & 0.78 / 0.88 & 7.61 / 5.52 & 11.66 / 9.22 & 3.02 / 1.98 & 19.81 / 16.78 &  -  \\
wGE \cite{wGE}  & 3.96 / 3.55 & 2.62 / 2.19 & 0.56 / 0.61 & 9.20 / 8.62 & 15.69 / 14.66 & 3.03 / 2.63 & 33.91 / 34.32 &  -  \\
PCA \cite{NUS}  & 4.33 / 3.90 & 3.16 / 2.41 & 0.54 / 0.57 & 10.02 / 9.49 & 16.75 / 15.87 & 3.46 / 2.94 & 32.40 / 32.84 &  - \\ 
MSGP \cite{MSGP}  & 6.41 / 5.88 & 5.48 / 4.01 & 0.84 / 1.03 & 14.20 / 13.16 & 23.99 / 23.26 & 5.38 / 4.59 & 34.23 / 35.95 &  - \\ 
GI \cite{GI}  & 4.30 / 4.50 & 2.78 / 2.74 & 0.43 / 0.75 & 11.14 / 11.11 & 21.42 / 20.61 & 2.94 / 3.13 & 32.52 / 32.96 &  - \\ 
TECC \cite{TECC}  & 3.78 / 3.30 & 2.69 / 2.23 & 0.62 / 0.59 & 8.49 / 7.66 & 14.16 / 13.18 & 2.99 / 2.57 & 31.41 / 31.75 &  - \\ 
Gamut (pixels) \cite{GAMUT}   & 3.72 / 2.54 & 2.83 / 1.61 & 0.73 / 0.61 & 8.10 / 5.99 & 13.66 / 10.52 & 3.01 / 1.83 & 21.81 / 17.82 &  0.636 \\ 
Gamut (edges) \cite{GAMUT}  & 4.43 / 3.92 & 3.34 / 3.04 & 1.04 / 1.13 & 9.52 / 8.13 & 15.13 / 13.50 & 3.62 / 3.22 & 19.26 / 15.91 &  324 \\
Gamut (1st) \cite{GAMUT}  & 4.33 / 3.85 & 3.34 / 2.58 & 0.68 / 0.98 & 9.87 / 8.77 & 15.61 / 13.86 & 3.52 / 2.83 & 22.15 / 19.03 &  279 \\
NIS \cite{NIS}  & 4.28 / 3.80 & 3.75 / 2.73 & 0.72 / 0.85 & 9.14 / 8.21 & 14.57 / 13.33 & 3.80 / 3.09 & 31.72 / 31.87 &  0.078 \\ 
Classification-CC \cite{CLASSIFICATION}  & 2.55 / 1.64 & 2.15 / 1.15 & 0.63 / 0.36 & 5.37 / 3.57 & 8.88 / 5.47 & 2.17 / 1.32 & 17.82 / 7.89 &   58,384\\ 
FFCC \cite{FFCC}  & 2.21 / 1.54 & 1.33 / 0.90 & 0.42 / \colorbox{secondbest}{0.24} & 5.46 / 3.92 & 10.38 / 8.03 & 1.55 / 1.00 & 17.18 / 14.53 &  12 \\ 
FFCC (capture info) \cite{FFCC}  & 1.97 / 1.43 & 1.25 / 0.80 & 0.39 / \colorbox{secondbest}{0.24} & 4.72 / 3.65 & 8.56 / 7.13 & 1.43 / 0.93 & 14.16 / 12.30 &  36.9 \\ 
FC4 \cite{FC4}  & 4.88 / 5.49 & 2.97 / 3.66 & 0.90 / 1.32 & 12.41 / 13.00 & 31.38 / 31.36 & 3.15 / 3.85 & 44.29 / 42.46 &  1,705 \\ 
APAP (GW) \cite{APAP}  & 3.43 / 1.99 & 2.66 / 1.53 & 0.86 / 0.52 & 7.25 / 4.11 & 10.91 / 6.02 & 2.92 / 1.66 & 15.11 / 7.10 &  0.289 \\ 
SIIE \cite{SIIE}  & 3.67 / - & 3.16 / - & 0.88 / - & 7.44 / - & 10.75 / - & 3.24 / - & 16.91 / - &  1,008 \\ 
SIIE (tuned) \cite{SIIE}  & 2.90 / - & 2.23 / - & 0.45 / - & 6.44 / - & 10.27 / - & 2.38 / - & 13.97 / - &  1,008 \\ 
SIIE (tuned-CS) \cite{SIIE}  & 2.65 / 1.61 & 1.91 / 1.27 & 0.45 / 0.32 & 6.13 / 3.58 & 10.76 / 5.34 & 2.05 / 1.35 & 20.53 / 8.12 &  1,008 \\ 
KNN (raw) \cite{knn}  & 2.43 / 1.42 & 1.52 / 0.99 & 0.34 / 0.25 & 6.08 / 3.27 & 11.70 / 5.90 & 1.63 / 1.06 & 22.00 / 8.96 &  757 \\
Quasi-U-CC \cite{QUASI-CC}  & 3.60 / 3.27 & 2.71 / 2.10 & 0.50 / 0.55 & 8.19 / 7.87 & 13.18 / 12.68 & 2.89 / 2.41 & 22.60 / 23.17 &  54,421 \\
Quasi-U-CC (tuned) \cite{QUASI-CC}  & 2.70 / 2.46 & 1.92 / 1.53 & 0.50 / 0.51 & 6.21 / 5.88 & 9.90 / 9.24 & 2.12 / 1.76 & 15.20 / 16.99 &  54,421 \\
BoCF \cite{BoCF}  & 3.12 / 1.94 & 2.55 / 1.37 & 0.85 / 0.50 & 6.28 / 4.19 & 8.91 / 6.90 & 2.67 / 1.57 & \colorbox{secondbest}{11.97} / 10.81 &  59 \\     
C4 \cite{C4} & \colorbox{best}{1.63} / 1.46 & \colorbox{best}{1.04} / 0.94 & 0.30 / 0.29 & \colorbox{secondbest}{3.87} / 3.49 & \colorbox{best}{6.73} / 5.68 & \colorbox{best}{1.17} / 1.07 & \colorbox{best}{9.89} / 10.59 &  5,116 \\ 
CWCC \cite{CWCC}  & 3.21 / 2.21 & 2.44 / 1.79 & 0.83 / 0.73 & 6.84 / 4.48 & 10.67 / 7.89 & 2.68 / 1.83 & 12.99 / 14.34 &  101 \\ 
C5 \cite{C5}  & 2.90 / - & 2.34 / - & 0.78 / - & 5.95 / - & 9.89 / - & 2.44 / - & 19.78 / - &  412 \\ 
C5 (tuned) \cite{C5}  & 1.87 / - & 1.14 / - & \colorbox{secondbest}{0.29} / - & 4.74 / - & 8.69 / - & 1.26 / - & 12.44 / - &  412 \\ 
C5 (tuned-CS) \cite{C5}  & 1.80 / 1.44 & 1.24 / 0.92 & 0.33 / 0.27 & 4.23 / 3.45 & 7.44 / 5.69 & 1.37 / 1.05 & 13.70 / 8.17 &  172 \\ 
TLCC \cite{TLCC}  & 2.69 / 2.51 & 2.09 / 1.77 & 0.63 / 0.57 & 5.75 / 5.60 & 9.22 / 9.63 & 2.21 / 1.98 & 13.51 / 21.24 &  32,910 \\ 
PCC \cite{PCC}  & 3.06 / 1.89 & 1.92 / 1.39 & 0.46 / 0.40 & 7.29 / 4.18 & 11.99 / 6.91 & 2.28 / 1.48 & 24.28 / 9.33 &  0.378 \\ 
RGP \cite{RGP}  & 4.31 / 4.39 & 2.92 / 2.81 & 0.39 / 0.69 & 10.84 / 10.82 & 20.28 / 18.68 & 3.06 / 3.26 & 33.93 / 34.36 &  - \\ 
CFCC \cite{CFCC}  & 2.74 / 1.57 & 2.04 / 1.25 & 0.57 / 0.42 & 6.17 / 3.34 & 10.45 / 5.82 & 2.18 / 1.29 & 14.53 / 9.35 &  0.283 \\ \hdashline
Ours (w/o $\mathbf{H}$, $h=0$)   & 2.24 / 1.60 & 1.68 / 1.20 & 0.47 / 0.38 & 4.93 / 3.61 & 8.83 / 6.18 & 1.77 / 1.25 & 19.44 / 8.52 &  2.1  \\
Ours (w/o $\mathbf{c}$)   & 2.28 / 1.35 & 1.51 / 0.89 & 0.35 / 0.29 & 5.64 / 3.12 & 12.01 / 5.86 & 1.61 / 0.93 & 23.51 / 10.53 &   4.07\\
Ours (w/o $\mathbf{H}$, $\mathbf{c}=\mathbf{p}$)  & 5.28 / 4.20 & 3.15 / 2.28 & 0.61 / 0.55 & 13.39 / 10.83 & 19.41 / 15.72 & 3.83 / 2.84 & 27.27 / 19.60 &  1.95 \\
Ours (w/o $\mathbf{H}$, $\mathbf{c}=\left[\mathbf{p}^T, i\right]^T$)  & 4.22 / 3.41 & 2.15 / 2.23 & 0.43 / 0.50 & 10.86 / 8.30 & 18.10 / 13.93 & 2.93 / 2.57 & 27.53 / 18.70 &   1.97 \\
Ours (w/o $\mathbf{H}$, $\mathbf{c}=\left[\mathbf{p}^T, s\right]^T$)  & 4.61 / 3.79 & 2.34 / 2.23 & 0.59 / 0.52 & 12.14 / 9.61 & 18.89 / 14.74 & 3.06 / 2.61 & 26.83 / 19.58 &  1.97 \\
Ours (w/o $\mathbf{H}$, $\mathbf{c}=\left[\mathbf{p}^T, f\right]^T$)  & 5.24 / 4.14 & 2.67 / 2.30 & 0.52 / 0.55 & 13.62 / 10.69 & 19.67 / 15.73 & 3.60 / 2.72 & 26.57 / 19.67 &  1.97 \\
Ours (w/o $\mathbf{H}$, $\mathbf{c}=\left[\mathbf{p}^T, \mathbf{n}^T\right]^T$)  & 2.34 / 1.66 & 1.66 / 1.14 & 0.41 / 0.38 & 5.50 / 3.83 & 9.74 / 6.89 & 1.73 / 1.27 & 19.47 / 9.22 &  2.05 \\
Ours ($\mathbf{c}=\left[i\right]^T$)  & 1.99 / 1.26 & 1.35 / 0.82 & 0.36 / 0.28 & 4.78 / 3.01 & 9.25 / 5.88 & 1.51 / 0.92 & 16.89 / 9.82 &  4.61 \\
Ours ($\mathbf{c}=\left[s\right]^T$)  & 2.11 / 1.27 & 1.57 / 0.81 & 0.38 / 0.25 & 4.90 / 3.10 & 9.19 / 5.98 & 1.61 / 0.89 & 20.73 / 9.65 &  4.61 \\
Ours ($\mathbf{c}=\left[f\right]^T$)  & 2.06 / 1.33 & 1.24 / 0.80 & 0.36 / \colorbox{secondbest}{0.24} & 5.11 / 3.34 & 10.34 / 6.04 & 1.38 / 0.91 & 20.66 / 11.15 &  4.61\\
Ours ($\mathbf{c}=\mathbf{n}$)  & 1.94 / 1.26 & 1.20 / 0.88 & 0.38 / 0.32 & 4.66 / 2.93 & 8.85 / 5.77 & 1.36 / 0.88 & 23.85 / 9.55 &   4.69\\
Ours ($\mathbf{c}=\mathbf{n}$)  & 1.93 / 1.28 & 1.27 / 0.83 & 0.36 / \colorbox{best}{0.23} & 4.71 / 3.26 & 9.36 / 5.96 & 1.35 / 0.86 & 20.59 / 9.76 &  4.79 \\
Ours (w/o $\mathbf{H}_{e}$)  & 1.96 / 1.30 & 1.37 / 0.87 & 0.35 / \colorbox{best}{0.23} & 4.73 / 3.03 & 10.29 / \colorbox{best}{4.65} & 1.41 / 0.96 & 26.11 / \colorbox{best}{5.84} &   4.86\\                    
Ours (w/o $u$/$v$ coord.)  & \colorbox{secondbest}{1.69} / 1.34 & 1.25 / 0.94 & \colorbox{best}{0.28} / 0.31 & 3.88 / 3.02 & \colorbox{secondbest}{7.03} / 4.89 & 1.32 / 1.06 & 19.58 / \colorbox{secondbest}{6.84} &  4.79 \\
Ours (w/ $log$-$\mathbf{H}$ \cite{CCC, C5})   & 1.88 / 1.27 & 1.37 / 0.89 & 0.40 / 0.26 & 4.24 / 2.86 & 7.28 / 4.92 & 1.43 / 1.00 & 23.83 / 6.96 &   4.93\\  
Ours ($\mathbf{H}$ $\rightarrow$ $\mathbf{I}_\texttt{chroma}$)  & 2.17 / 1.32 & 1.46 / 0.85 & 0.41 / 0.25 & 5.20 / 3.19 & 10.29 / 5.65 & 1.58 / 0.93 & 17.97 / 7.61 &  4.79 \\
Ours (w/ $\mathbf{c}$-raw)  & 2.11 / 1.46 & 1.45 / 1.10 & 0.46 / 0.38 & 4.79 / 3.23 & 9.00 / 5.51 & 1.60 / 1.17 & 21.59 / 10.09 &  4.93 \\
Ours ($h=24$)  & 1.74 / 1.14 & 1.24 / 0.74 & \colorbox{secondbest}{0.29} / \colorbox{best}{0.23} & 4.12 / 2.74 & 7.80 / \colorbox{secondbest}{4.84} & 1.32 / 0.84 & 21.94 / 7.24 &  4.93 \\
Ours ($\mathbf{I} \in \mathbb{R}^{(64\times48)\times3}$)  & 1.86 / 1.16 & 1.26 / \colorbox{secondbest}{0.73} & 0.34 / \colorbox{secondbest}{0.24} & 4.35 / 2.81 & 8.36 / 4.91 & 1.41 / \colorbox{secondbest}{0.83} & 18.76 / 8.33 &  4.93 \\

Ours (w/o $\mathbf{p}$) & 1.83 / 1.13 & 1.25 / 0.75 & 0.31 / 0.20 & 4.39 / 2.72 & 8.87 / 4.90 & 1.34 / 0.83 & 18.52 / 7.33 & 4.83 \\
Ours (w/o $\mathbf{n}$, w/o $\mathbf{r}$) & 1.79 / 1.24 & 1.20 / 0.86 & 0.34 / \colorbox{secondbest}{0.24} & 4.23 / 2.95 & 7.77 / 5.53 & 1.35 / 0.89 & 17.82 / 10.33 & 4.83 \\
Ours (w/o $\mathbf{n}$, w/ $\mathbf{r}$)   & 1.70 / \colorbox{secondbest}{1.12} & 1.15 / 0.85 & \colorbox{secondbest}{0.29} / 0.25 & 4.13 / \colorbox{best}{2.56} & 8.84 / \colorbox{best}{4.65} & 1.23 / 0.87 & 21.55 / 7.79 &  4.93 \\
Ours (w/ $\mathbf{n}$, w/ $\mathbf{r}$)  & \colorbox{best}{1.63} / \colorbox{secondbest}{1.12} & 1.14 / \colorbox{best}{0.71} & 0.33 / 0.25 & \colorbox{best}{3.78} / 2.67 & 7.19 / 4.98 & 1.25 / \colorbox{secondbest}{0.83} & 19.96 / 9.28 &  5.03 \\
Ours (w/ $\mathbf{n}$, w/o $\mathbf{r}$)  & \colorbox{best}{1.63} / \colorbox{best}{1.09} & \colorbox{secondbest}{1.05} / \colorbox{best}{0.71} & \colorbox{secondbest}{0.29} / \colorbox{secondbest}{0.24} & 3.92 / \colorbox{secondbest}{2.62} & 8.12 / 4.89 & \colorbox{secondbest}{1.18} / \colorbox{best}{0.77} & 22.22 / 8.45 &  4.93 \\

\end{tabular}
}
\end{table*}

%% file: tables/results_outdoor_indoor_test_wo_mask.tex
\begin{table*}[!t]
    \centering
    \caption{\textbf{Results on outdoor vs. indoor scenes}. We report the mean, median, best 25\%, worst 25\%, tri-mean, and maximum angular errors for each experiment setting on the testing set (without masking). Models are trained and tested on the neutral ground-truth illuminants. $\mathbf{c}$ denotes the time-capture feature. $\mathbf{p}$ represents the time feature. `$\mathbf{c}=\text{all}$' indicates that the full time-capture feature is used. \label{tab:results-outdoor-indoor}}\vspace{-1mm}
    \scalebox{0.75}{
     \addtolength{\tabcolsep}{-0.2em}
    \begin{tabular}{l|cc|cc|cc|cc|cc|cc|cc}
    \multirow{2}{*}{\textbf{Method}} & \multicolumn{2}{c|}{\textbf{Mean}} & \multicolumn{2}{c|}{\textbf{Med.}} & \multicolumn{2}{c|}{\textbf{Best 25\%}} & \multicolumn{2}{c|}{\textbf{Worst 25\%}} & \multicolumn{2}{c|}{\textbf{Worst 5\%}} & \multicolumn{2}{c|}{\textbf{Tri.}} & \multicolumn{2}{c}{\textbf{Max}} \\ \cline{2-15} 
                            & outdoor & indoor & outdoor & indoor & outdoor & indoor & outdoor & indoor & outdoor & indoor & outdoor & indoor & outdoor & indoor \\ \hline
    Ours (w/o $\mathbf{H}$, $\mathbf{c}=\mathbf{p}$)  & 3.47 & 9.39 & 1.96 & 8.83 & 0.40 & 1.98 & 9.24 & 17.73 & 16.19 & 25.51 & 2.28 & 8.97 & 24.31 & 36.23 \\ \hline
    Ours (w/o $\mathbf{H}$, $\mathbf{c}=\text{all}$)  & 2.12 & 3.05 & 1.41 & 2.41 & 0.38 & 1.13 & 5.05 & 6.00 & 8.58 & 10.00 & 1.53 & 2.53 & 17.55 & 11.25   \\ \hline
    Ours (w/ $\mathbf{H}$, $\mathbf{c}=\text{all}$)   & 1.77 & 1.97 & 1.20 & 1.26 & 0.33 & 0.42 & 4.18 & 4.83 & 8.32 & 10.15 & 1.31 & 1.33 & 18.75 & 35.42         
\end{tabular}
}
\end{table*}

%% file: tables/results_test_w_mask.tex
\begin{table*}[t]
\centering
\caption{\textbf{Results on the testing set with masking}. We report the mean, median, best 25\%, worst 25\%, tri-mean, and maximum angular errors for each method on both neutral and user-preference white-balance ground truth, presented in the format (neutral / user-preference).  Symbols $\mathbf{n}$ and $\mathbf{r}$ represent noise stats and SNR stats, respectively. The \colorbox{best}{best} and \colorbox{secondbest}{second-best} results are highlighted. \label{tab:results-test}}\vspace{-1mm}
\scalebox{0.8}{
\begin{tabular}{l|ccccccc}
\textbf{Method} & \textbf{Mean}  & \textbf{Med.}  & \begin{tabular}[c]{@{}c@{}}\textbf{Best} \\ \textbf{25\%}\end{tabular} & \begin{tabular}[c]{@{}c@{}}\textbf{Worst} \\ \textbf{25\%}\end{tabular} & \begin{tabular}[c]{@{}c@{}}\textbf{Worst} \\ \textbf{5\%}\end{tabular} & \textbf{Tri.}   & \textbf{Max} \\ \hline

GW \cite{GW}   & 6.38 / 5.68 & 6.29 / 4.67 & 1.05 / 1.11 & 12.55 / 12.05 & 18.15 / 19.89 & 5.94 / 4.86 & 28.09 / 31.89 \\ 
SoG \cite{SoG}   & 4.36 / 3.78 & 3.44 / 2.19 & 0.66 / 0.70 & 9.52 / 9.27 & 14.53 / 16.36 & 3.68 / 2.69 & 23.22 / 32.06 \\ 
GE-1st \cite{GE}   & 4.02 / 3.58 & 3.13 / 2.27 & 0.63 / 0.59 & 9.11 / 8.86 & 14.37 / 16.09 & 3.32 / 2.59 & 21.38 / 32.21 \\ 
GE-2nd \cite{GE}   & 3.90 / 3.38 & 2.87 / 1.97 & 0.65 / 0.59 & 8.82 / 8.39 & 14.02 / 15.65 & 3.11 / 2.32 & 22.24 / 31.73 \\ 
Max-RGB \cite{maxRGB}  & 3.70 / 2.73 & 2.80 / 1.87 & 0.90 / 0.91 & 7.90 / 6.19 & 12.32 / 12.76 & 3.05 / 1.95 & 21.84 / 24.86 \\ 
wGE \cite{wGE}   & 3.76 / 3.33 & 2.68 / 2.01 & 0.57 / 0.53 & 8.66 / 8.45 & 13.85 / 16.01 & 2.98 / 2.33 & 21.39 / 31.94 \\ 
PCA \cite{NUS}   & 4.34 / 3.75 & 3.52 / 1.97 & 0.62 / 0.64 & 9.54 / 9.47 & 14.51 / 16.65 & 3.67 / 2.58 & 22.87 / 32.31 \\ 
MSGP \cite{MSGP}   & 6.62 / 5.87 & 5.91 / 4.56 & 0.98 / 1.05 & 13.56 / 12.87 & 20.62 / 22.30 & 5.85 / 4.85 & 37.25 / 36.92 \\ 
GI \cite{GI}   & 4.76 / 4.85 & 3.24 / 2.83 & 0.45 / 0.76 & 11.65 / 12.07 & 19.87 / 21.06 & 3.52 / 3.46 & 36.34 / 36.02 \\ 
TECC \cite{TECC}   & 3.92 / 3.39 & 2.95 / 2.11 & 0.67 / 0.58 & 8.85 / 8.41 & 13.68 / 15.43 & 3.21 / 2.46 & 21.85 / 31.83 \\ 
Gamut (pixels) \cite{GAMUT}   & 3.53 / 2.53 & 2.53 / 1.46 & 0.69 / 0.59 & 7.95 / 6.39 & 12.35 / 12.86 & 2.88 / 1.65 & 21.53 / 26.62 \\ 
Gamut (edges) \cite{GAMUT}   & 4.28 / 4.06 & 3.30 / 3.03 & 1.07 / 0.96 & 9.23 / 9.07 & 14.47 / 16.86 & 3.48 / 3.18 & 28.50 / 30.72 \\ 
Gamut (1st) \cite{GAMUT}   & 3.87 / 3.81 & 2.80 / 2.51 & 0.70 / 0.85 & 8.88 / 8.83 & 14.01 / 16.17 & 3.03 / 2.86 & 21.38 / 32.62 \\ 
NIS \cite{NIS}   & 4.53 / 4.03 & 3.69 / 2.74 & 0.67 / 0.78 & 9.80 / 9.40 & 14.83 / 16.12 & 3.77 / 3.10 & 20.76 / 32.75 \\ 
Classification-CC \cite{CLASSIFICATION}   & 2.71 / 1.68 & 2.07 / 1.19 & 0.60 / 0.35 & 5.94 / 3.78 & 9.83 / 6.28 & 2.21 / 1.32 & 19.31 / 10.26 \\ 
FFCC \cite{FFCC}   & 2.61 / 1.54 & 1.43 / 0.85 & 0.37 / 0.26 & 6.83 / 4.07 & 16.24 / 8.46 & 1.66 / 0.98 & 48.98 / 18.60 \\ 
FFCC (capture info) \cite{FFCC}   & 2.19 / 1.37 & 1.37 / 0.82 & \colorbox{best}{0.30} / 0.24 & 5.49 / 3.53 & 11.86 / 6.90 & 1.53 / 0.92 & 48.53 / 16.40 \\ 
FC4 \cite{FC4}  & 3.92 / 2.67 & 2.77 / 2.23 & 0.84 / 0.89 & 9.17 / 5.18 & 17.30 / 7.77 & 2.88 / 2.36 & 45.83 / 11.83 \\ 
APAP (GW) \cite{APAP}  & 3.77 / 2.13 & 3.20 / 1.70 & 0.98 / 0.48 & 7.63 / 4.49 & 10.96 / 6.69 & 3.34 / 1.80 & \colorbox{secondbest}{14.66} / \colorbox{best}{8.91} \\ 
SIIE \cite{SIIE}  & 4.16 / - & 3.37 / - & 0.91 / - & 9.06 / - & 16.13 / - & 3.49 / - & 43.76 / - \\ 
SIIE (tuned) \cite{SIIE}  & 3.16 / - & 2.25 / - & 0.49 / - & 7.27 / - & 12.25 / - & 2.50 / - & 34.53 / - \\ 
SIIE (tuned-CS) \cite{SIIE}  & 3.17 / 1.79 & 2.20 / 1.21 & 0.49 / 0.32 & 7.50 / 4.23 & 13.76 / 6.94 & 2.41 / 1.34 & 39.01 / 9.54 \\ 
KNN (raw) \cite{knn}   & 2.41 / 1.44 & 1.50 / 0.85 & \colorbox{secondbest}{0.34} / \colorbox{secondbest}{0.21} & 6.14 / 3.70 & 12.05 / 7.12 & 1.65 / 1.00 & 28.95 / 13.49 \\ 
Quasi-U-CC \cite{QUASI-CC}   & 3.84 / 3.38 & 2.94 / 1.89 & 0.55 / 0.60 & 8.55 / 8.56 & 13.62 / 16.02 & 3.20 / 2.36 & 24.74 / 32.79 \\ 
Quasi-U-CC (tuned) \cite{QUASI-CC}   & 3.02 / 2.70 & 2.18 / 1.51 & 0.48 / 0.48 & 6.95 / 6.94 & 11.38 / 14.05 & 2.37 / 1.74 & 22.67 / 33.64 \\ 
BoCF \cite{BoCF}   & 3.44 / 2.14 & 2.67 / 1.60 & 0.89 / 0.49 & 7.16 / 4.72 & 11.07 / 7.84 & 2.89 / 1.71 & 21.38 / 19.68 \\ 
C4 \cite{C4}   & \colorbox{best}{1.73} / 1.45 & \colorbox{best}{1.18} / 0.90 & 0.35 / 0.24 & \colorbox{best}{4.09} / 3.67 & \colorbox{secondbest}{7.60} / 7.05 & \colorbox{secondbest}{1.29} / 1.00 & 21.55 / 18.09 \\ 
CWCC \cite{CWCC}   & 3.46 / 2.31 & 2.48 / 1.71 & 0.76 / 0.70 & 7.62 / 4.98 & 11.84 / 9.54 & 2.82 / 1.85 & 20.73 / 20.81 \\ 
C5 \cite{C5}   & 3.18 / - & 2.49 / - & 0.78 / - & 6.82 / - & 10.55 / - & 2.67 / - & 16.70 / - \\ 
C5 (tuned-CS) \cite{C5}   & \colorbox{secondbest}{1.81} / 1.27 & 1.22 / 0.86 & 0.36 / 0.22 & \colorbox{secondbest}{4.31} / \colorbox{secondbest}{2.98} & \colorbox{best}{7.45} / \colorbox{best}{5.08} & 1.36 / 0.95 & 16.78 / \colorbox{secondbest}{9.23} \\ 
TLCC \cite{TLCC}   & 2.64 / 2.87 & 2.06 / 2.01 & 0.65 / 0.69 & 5.69 / 6.58 & 9.68 / 13.13 & 2.16 / 2.18 & 21.44 / 33.20 \\ 
PCC \cite{PCC}  & 2.87 / 1.68 & 2.03 / 1.16 & 0.51 / 0.38 & 6.83 / 3.89 & 10.62 / 7.05 & 2.25 / 1.25 & \colorbox{best}{14.60} / 10.34 \\ 
RGP \cite{RGP}   & 4.57 / 4.50 & 3.21 / 2.88 & 0.43 / 0.67 & 10.95 / 11.16 & 18.06 / 20.05 & 3.55 / 3.33 & 32.11 / 33.98 \\ 
CFCC \cite{CFCC}   & 2.82 / 1.50 & 2.01 / 1.02 & 0.70 / 0.38 & 6.24 / 3.46 & 10.62 / 6.39 & 2.19 / 1.10 & 17.11 / 10.19 \\ \hdashline
Ours (w/o $\mathbf{n}$, w/o $\mathbf{r}$) & 1.86 / 1.26 & 1.31 / \colorbox{secondbest}{0.78} & 0.37 / 0.23 & 4.33 / 3.15 & 8.71 / 6.00 & 1.41 / 0.89 & 22.63 / 16.07 \\
Ours (w/o $\mathbf{n}$, w/ $\mathbf{r}$)  & 1.84 / 1.25 & \colorbox{best}{1.18} / 0.80 & \colorbox{secondbest}{0.34} / 0.24 & 4.55 / 3.04 & 9.36 / \colorbox{secondbest}{5.15} & 1.31 / 0.90 & 24.99 / 13.02 \\ 
Ours (w/ $\mathbf{n}$, w/o $\mathbf{r}$)   & 1.84 / \colorbox{secondbest}{1.22} & \colorbox{best}{1.18} / \colorbox{best}{0.73} & 0.35 / 0.24 & 4.56 / 3.06 & 9.85 / 5.76 & \colorbox{best}{1.25} / \colorbox{best}{0.82} & 29.46 / 15.24 \\ 
Ours (w/ $\mathbf{n}$, w/ $\mathbf{r}$)   & 1.82 / \colorbox{best}{1.21} & \colorbox{secondbest}{1.19} / \colorbox{secondbest}{0.78} & 0.36 / \colorbox{best}{0.20} & 4.42 / \colorbox{best}{2.97} & 9.42 / 5.18 & \colorbox{best}{1.25} / \colorbox{secondbest}{0.87} & 35.42 / 12.93 \\ 
                                                                                                      
\end{tabular}
}
\end{table*}

%% file: tables/results_val_wo_mask.tex
\begin{table*}[t]
\centering
\caption{\textbf{Results on the validation set without masking}. We report the mean, median, best 25\%, worst 25\%, tri-mean, and maximum angular errors for each method on neutral and user-preference white-balance ground-truth illuminants, presented in the format (neutral / user-preference).  Symbols $\mathbf{n}$ and $\mathbf{r}$ refer to noise and SNR stats, respectively. The \colorbox{best}{best} and \colorbox{secondbest}{second-best} results are highlighted. \label{tab:results-validation-wo-mask}}\vspace{-1mm}
\scalebox{0.8}{
\begin{tabular}{l|ccccccc}
\textbf{Method} & \textbf{Mean}  & \textbf{Med.}  & \begin{tabular}[c]{@{}c@{}}\textbf{Best} \\ \textbf{25\%}\end{tabular} & \begin{tabular}[c]{@{}c@{}}\textbf{Worst} \\ \textbf{25\%}\end{tabular} & \begin{tabular}[c]{@{}c@{}}\textbf{Worst} \\ \textbf{5\%}\end{tabular} & \textbf{Tri.}   & \textbf{Max} \\ \hline

GW \cite{GW}   & 5.76 / 5.30 & 4.83 / 3.94 & 0.78 / 0.89 & 12.45 / 11.68 & 19.83 / 18.98 & 4.91 / 4.29 & 30.03 / 30.51 \\ 
SoG \cite{SoG}   & 4.42 / 3.74 & 3.27 / 2.35 & 0.59 / 0.71 & 10.01 / 8.76 & 15.91 / 14.41 & 3.65 / 2.77 & 30.43 / 30.82 \\ 
GE-1st \cite{GE}   & 4.15 / 3.40 & 2.99 / 2.30 & 0.63 / 0.61 & 9.42 / 8.13 & 16.51 / 14.63 & 3.28 / 2.61 & 32.02 / 32.37 \\ 
GE-2nd \cite{GE}   & 3.88 / 3.10 & 2.74 / 2.01 & 0.69 / 0.58 & 8.68 / 7.27 & 15.31 / 13.45 & 3.06 / 2.29 & 28.72 / 29.15 \\ 
Max-RGB \cite{maxRGB}  & 3.76 / 2.57 & 2.98 / 1.79 & 0.96 / 0.92 & 7.94 / 5.56 & 13.21 / 10.13 & 3.17 / 1.91 & 21.06 / 17.57 \\ 
wGE \cite{wGE}   & 4.11 / 3.28 & 2.65 / 2.06 & 0.59 / 0.53 & 9.66 / 8.30 & 17.98 / 15.37 & 3.00 / 2.33 & 33.88 / 34.29 \\ 
PCA \cite{NUS}   & 4.36 / 3.77 & 3.04 / 2.15 & 0.56 / 0.54 & 10.27 / 9.32 & 17.99 / 16.41 & 3.37 / 2.69 & 32.27 / 32.71 \\ 
MSGP \cite{MSGP}   & 6.39 / 5.81 & 5.48 / 3.87 & 0.80 / 0.99 & 14.08 / 13.17 & 23.97 / 23.36 & 5.35 / 4.51 & 34.23 / 35.95 \\ 
GI \cite{GI}   & 4.21 / 4.53 & 2.75 / 2.79 & 0.42 / 0.76 & 10.82 / 11.08 & 21.20 / 20.40 & 2.85 / 3.17 & 32.52 / 32.96 \\ 
TECC \cite{TECC}   & 3.89 / 3.08 & 2.80 / 2.00 & 0.64 / 0.56 & 8.80 / 7.35 & 15.47 / 13.64 & 3.03 / 2.26 & 28.28 / 28.69 \\ 
Gamut (pixels) \cite{GAMUT}   & 3.72 / 2.54 & 2.83 / 1.61 & 0.73 / 0.61 & 8.10 / 5.99 & 13.66 / 10.52 & 3.01 / 1.83 & 21.81 / 17.82 \\ 
Gamut (edges) \cite{GAMUT}   & 4.42 / 3.92 & 3.37 / 3.09 & 1.05 / 1.13 & 9.48 / 8.12 & 15.14 / 13.49 & 3.65 / 3.25 & 19.24 / 15.91 \\ 
Gamut (1st) \cite{GAMUT}   & 4.33 / 3.85 & 3.34 / 2.58 & 0.68 / 0.98 & 9.87 / 8.77 & 15.61 / 13.86 & 3.52 / 2.83 & 22.16 / 19.03 \\ 
NIS \cite{NIS}   & 4.36 / 3.80 & 3.44 / 2.73 & 0.80 / 0.85 & 9.35 / 8.21 & 14.88 / 13.33 & 3.70 / 3.09 & 31.43 / 31.87 \\ 
Classification-CC \cite{CLASSIFICATION}   & 2.58 / 1.61 & 2.25 / 1.23 & 0.58 / 0.35 & 5.32 / 3.53 & 9.23 / 5.33 & 2.24 / 1.29 & 18.53 / \colorbox{secondbest}{6.95} \\ 
FFCC \cite{FFCC}   & 2.19 / 1.51 & 1.29 / 0.91 & 0.42 / \colorbox{secondbest}{0.25} & 5.42 / 3.76 & 10.20 / 7.70 & 1.53 / 1.01 & 17.18 / 13.79 \\ 
FFCC (capture info) \cite{FFCC}   & 1.97 / 1.35 & 1.29 / 0.91 & 0.35 / \colorbox{secondbest}{0.25} & 4.70 / 3.22 & 8.23 / 5.59 & 1.47 / 1.02 & 16.01 / 8.74 \\ 
FC4 \cite{FC4}   & 4.02 / 2.87 & 2.92 / 2.72 & 0.90 / 0.83 & 9.26 / 5.24 & 19.13 / 7.23 & 2.98 / 2.72 & 39.64 / 11.83 \\ 
APAP (GW) \cite{APAP}   & 3.38 / 1.93 & 2.54 / 1.52 & 0.86 / 0.51 & 7.15 / 4.02 & 11.04 / 5.96 & 2.80 / 1.58 & 16.12 / 7.16 \\ 
SIIE \cite{SIIE}  & 3.67 / - & 3.16 / - & 0.88 / - & 7.44 / - & 10.75 / - & 3.24 / - & 16.91 / - \\ 
SIIE (tuned) \cite{SIIE}  & 2.90 / - & 2.23 / - & 0.45 / - & 6.44 / - & 10.27 / - & 2.38 / - & 13.97 / - \\ 
SIIE (tuned-CS) \cite{SIIE}  & 2.65 / 1.61 & 1.91 / 1.27 & 0.45 / 0.32 & 6.13 / 3.58 & 10.76 / 5.34 & 2.05 / 1.35 & 20.53 / 8.12 \\ 
KNN (raw) \cite{knn}   & 2.49 / 1.39 & 1.61 / 0.99 & 0.35 / 0.26 & 6.13 / 3.18 & 11.35 / 5.70 & 1.72 / 1.05 & 20.63 / 7.91 \\ 
Quasi-U-CC \cite{QUASI-CC}   & 3.66 / 3.19 & 2.61 / 1.95 & 0.55 / 0.53 & 8.58 / 7.79 & 14.24 / 12.77 & 2.84 / 2.29 & 22.69 / 23.25 \\ 
Quasi-U-CC (tuned) \cite{QUASI-CC}   & 2.82 / 2.36 & 1.99 / 1.51 & 0.55 / 0.47 & 6.46 / 5.60 & 9.93 / 8.77 & 2.21 / 1.70 & \colorbox{best}{11.32} / 11.99 \\ 
BoCF \cite{BoCF}   & 3.18 / 1.97 & 2.49 / 1.42 & 0.82 / 0.50 & 6.53 / 4.20 & 9.29 / 6.70 & 2.68 / 1.61 & \colorbox{secondbest}{12.17} / 10.60 \\ 
C4 \cite{C4}   & 1.72 / 1.42 & \colorbox{best}{1.04} / 0.86 & \colorbox{secondbest}{0.30} / 0.26 & 4.22 / 3.49 & \colorbox{secondbest}{7.19} / 5.44 & \colorbox{best}{1.21} / 1.03 & 14.36 / \colorbox{best}{6.49} \\ 
CWCC \cite{CWCC}   & 3.42 / 2.27 & 2.71 / 1.75 & 0.89 / 0.73 & 7.28 / 4.69 & 11.93 / 8.36 & 2.90 / 1.82 & 17.99 / 13.95 \\ 
C5 \cite{C5}   & 2.97 / - & 2.27 / - & 0.81 / - & 6.21 / - & 10.13 / - & 2.42 / - & 18.38 / - \\ 
C5 (tuned) \cite{C5}   & 2.00 / - & 1.21 / - & 0.31 / - & 5.09 / - & 9.71 / - & 1.31 / - & 17.56 / - \\ 
C5 (tuned-CS) \cite{C5}   & 2.01 / 1.46 & 1.36 / 0.96 & 0.36 / 0.26 & 4.82 / 3.53 & 8.82 / 5.68 & 1.49 / 1.06 & 15.99 / 8.17 \\ 
TLCC \cite{TLCC}   & 2.70 / 2.36 & 2.24 / 1.69 & 0.69 / 0.56 & 5.54 / 5.19 & 8.87 / 8.24 & 2.30 / 1.91 & 14.35 / 11.87 \\ 
PCC \cite{PCC}    & 3.32 / 1.90 & 2.19 / 1.34 & 0.48 / 0.39 & 7.89 / 4.33 & 13.12 / 7.49 & 2.52 / 1.45 & 24.28 / 11.30 \\ 
RGP \cite{RGP}  & 4.28 / 4.38 & 2.82 / 2.76 & 0.37 / 0.72 & 10.79 / 10.62 & 20.02 / 18.62 & 3.04 / 3.24 & 33.76 / 34.19 \\ 
CFCC \cite{CFCC}  & 2.98 / 1.70 & 2.13 / 1.18 & 0.62 / 0.43 & 6.83 / 3.90 & 11.92 / 7.85 & 2.33 / 1.28 & 19.49 / 14.06 \\ \hdashline

Ours (w/o $\mathbf{n}$, w/o $\mathbf{r}$) & 1.85 / 1.27 & 1.32 / 0.90 & 0.35 / 0.25 & 4.26 / 3.01 & 7.53 / 5.64 & 1.44 / 0.95 & 17.44 / 10.14 \\
Ours (w/o $\mathbf{n}$, w/ $\mathbf{r}$)   & 1.72 / \colorbox{best}{1.11} & 1.16 / 0.83 & \colorbox{secondbest}{0.30} / \colorbox{best}{0.24} & 4.13 / \colorbox{best}{2.58} & 8.16 / \colorbox{best}{4.66} & \colorbox{secondbest}{1.23} / \colorbox{secondbest}{0.86} & 18.50 / 7.54 \\ 
Ours (w/ $\mathbf{n}$, w/o $\mathbf{r}$)   & \colorbox{secondbest}{1.67} / \colorbox{best}{1.11} & \colorbox{secondbest}{1.07} / \colorbox{best}{0.70} & \colorbox{best}{0.29} / \colorbox{best}{0.24} & \colorbox{secondbest}{4.04} / \colorbox{secondbest}{2.68} & 7.90 / 4.97 & \colorbox{best}{1.21} / \colorbox{best}{0.76} & 24.40 / 8.28 \\ 
Ours (w/ $\mathbf{n}$, w/ $\mathbf{r}$)   & \colorbox{best}{1.66} / \colorbox{secondbest}{1.14} & 1.20 / \colorbox{secondbest}{0.73} & 0.33 / 0.26 & \colorbox{best}{3.77} / 2.69 & \colorbox{best}{6.95} / \colorbox{secondbest}{4.70} & 1.29 / \colorbox{secondbest}{0.86} & 19.80 / 6.98 \\ 

\end{tabular}
}
\end{table*}

%% file: tables/cross_cam.tex
\begin{table*}[t]
\centering
\caption{\textbf{Results on cross-camera generalization.} We report the mean angular error for each method on 257 test scenes captured using the S25 Ultra telephoto camera. None of the listed methods were trained on any data from the test camera. For our method, we present results for the model trained on data from the S24 Ultra main camera under three settings: without calibration, with \textit{offline} calibration, and with \textit{online} calibration. The \colorbox{best}{best} result is highlighted. \label{tab:cross-camera}}\vspace{-1mm}

\scalebox{0.7}{
\begin{tabular}{l|cc:ccc}
\multirow{2}{*}{\textbf{Method}} & \multirow{2}{*}{\textbf{C4 \cite{C4}}} & \multirow{2}{*}{\textbf{C5 \cite{C5}}} & \multicolumn{3}{c}{\textbf{Ours}}                                                                                                                       \\
                                 &                                                         &                                                         & \multicolumn{1}{l}{\textbf{w/o calibration}} & \multicolumn{1}{l}{\textbf{w/ offline calibration}} & \multicolumn{1}{l}{\textbf{w/ online calibration}} \\ \hline
\#params (K)                     & 5,116                                                   & 172                                                     & 4.8                                          & 4.8                                                 & 4.8                                                \\
Mean AE (S25-T)                  & 1.61                                                    & 1.63                                                    & 2.06                                         & 1.70                                                & \colorbox{best}{1.53}                                              
\end{tabular}
}
\end{table*}